# The Compatibility between the Pangu Weather Forecasting Model and Meteorological Operational Data

Wencong Cheng[1], Yan Yan[1], Jiangjiang Xia[2], Qi Liu[2], Chang Qu[2], Zhigang Wang[1]

[1]Beijing Aviation Meteorological Institute, Beijing 100085, China; emailtocheng@163.com, yanyanflykite@163.com, wzg_wang@163.com

[2]Key Laboratory of Regional Climate-Environment for Temperate East Asia & Center for Artificial Intelligence in Atmospheric Science, Institute of Atmospheric Physics, Chinese Academy of Sciences, Beijing 100029, China; xiajj@tea.ac.cn, liuqi@tea.ac.cn, quchang@tea.ac.cn

**Abstract:** Recently, multiple data-driven models based on machine learning for weather forecasting have emerged. These models are highly competitive in terms of accuracy compared to traditional numerical weather prediction (NWP) systems. In particular, the Pangu-Weather model, which is open source for non-commercial use, has been validated for its forecasting performance by the European Centre for Medium-Range Weather Forecasts (ECMWF) and has recently been published in the journal "Nature". In this paper, we evaluate the compatibility of the Pangu-Weather model with several commonly used NWP operational analyses through case studies. The results indicate that the Pangu-Weather model is compatible with different operational analyses from various NWP systems as the model initial conditions, and it exhibits a relatively stable forecasting capability. Furthermore, we have verified that improving the quality of global or local initial conditions significantly contributes to enhancing the forecasting performance of the Pangu-Weather model.

# 1 Introduction

With the rapid development of large artificial intelligence models like ChatGPTs, multiple Machine learning-based Weather forecasting Models (MWMs) with millions to billions of parameters have emerged, including the GNN model (Keisler, 2022), NVIDIA's "FourCastNet" model (Pathak *et al.*, 2022), Alibaba's "SwinVRNN" model (Hu *et al.*, 2023), Huawei's "Pangu-Weather" model (Bi *et al.*, 2023), DeepMind's "GraphCast" model (Lam *et al*., 2022), Shanghai AI Laboratory's "FengWu" model (Chen *et al.*, 2023a), and Fudan University's "FuXi" model (Chen *et al.*, 2023b). The forecast results of these models, according to their claims in the papers, have reached or exceeded the performance of the products from the European Centre for Medium-Range Weather Forecasts (ECMWF), leading to widespread attention in the meteorological community. Specifically, the forecasting skill of the Pangu-Weather model has been validated by the ECMWF, with the work recently published in the journal "Nature". Currently, the model's inference part has been open-sourced and is now available for testing and verification. By now, the Pangu-Weather Model is the only known large data-driven weather forecasting model with an open-sourced inference part.

While the forecasting performances of such MWMs have been validated through experiments conducted by the authors of these papers and some institutions, there are still some uncertainties that remain.

These uncertainties mainly include:

(1) Most of these MWMs utilized ERA5 reanalysis data as training datasets, and the initial conditions for model forecasts also employed ERA5 reanalysis data. While the ECMWF has recently evaluated the forecasting skill of the Pangu-Weather model using the operational analysis of Integrated Forecasting System (IFS) as input (Bouallègue, *et al.*, 2023), there are still other global medium-range NWP models' forecast products available. It remains to be experimentally validated whether the initial operational analyses of these NWP products can be used as the model inputs for these pretrained data-driven models, and whether they can achieve comparable forecasting performance compared to the corresponding NWP products. In other words, it is still unknown whether the forecasting performance of the Pangu-Weather model are limited to specific input data (such as ERA5 reanalysis data or operational analysis data of ECMWF-IFS related to the training dataset), or if they possess a certain level of generalizability and can be applied to various operational analyses of different NWP systems. Therefore, it is necessary to conduct experiments to verify the compatibility of the date-driven models to other NWP systems.

(2) Pangu-Weather model utilizes the whole global meteorological data (i.e. global NWP products) as input. However, due to constraints such as computational resources and data assimilation conditions, there are limited institutions capable of producing global NWP products by assimilating global data. Many institutions only have the ability to generate numerous local NWP products by assimilating local data. How to utilize the high-quality operational analyses of these local NWP systems to improve the quality of Pangu-Weather forecast products has not been analyzed.

Based on the aforementioned considerations, this study adopts the open-sourced Pangu-Weather model as the subject of experimental analysis. By means of case studies, besides the ERA5 reanalysis data used in these papers, four operational analyses of NWP products, which are utilized in various meteorological institutions for operational purposes, are employed as Pangu-Weather model's inputs. By using the Pangu-Weather model, the forecast products for the next 240 hours are generated. These products are then compared with corresponding NWP products to evaluate the generalizability of the Pangu-Weather model. We hope to promote the practical applications of data-driven weather forecasting models through this work.

## 2  Overview of the data-driven weather forecasting models.

The development of data-driven models based on machine learning (ML) for weather forecasting can be generally divided into four stages.

The first stage is the stage of exploratory research. The research in this stage mainly focuses on investigating whether neural network-based weather forecasting models have the potential to generate valid weather forecasts. For example, Dueben and Bauer (2018) developed a demonstration system that used 500hPa geopotential height and 2-meter temperature as forecasting targets, and verified that neural network methods can achieve forecast performance comparable to traditional NWP at coarse resolutions. These exploratory research works also includes studies by Scher (2018), Weyn *et al.* (2019), Scher and Messori (2019), and others.

The second stage is the establishment of standards for ML-based weather forecasting models. Rasp *et al.* (2020) constructed a standardized dataset using ERA5 reanalysis data, which can be used for training and testing MWMs. They also set some forecast levels from coarse-resolution NWP models and preliminary evaluation criteria. The aim is to enable future MWMs to be compared with each other under the same dataset and evaluation criteria, promoting the development of MWMs. Additionally, their work, along with Weyn *et al.* (2020), also presented the performance of using convolutional neural network (CNN) for forecasting, particularly in the case of Weyn *et al.* (2020) utilizing a deep learning UNet model and applying a cubed-sphere mapping to the data, resulting in better forecast results (Z500, 2m temperature, etc.) than those of the coarse-resolution model IFS T42.

The third stage is the breakthrough research stage of MWMs. Various institutions have developed large weather forecasting models (Keisler, 2022; Pathak *et al*., 2022; Hu *et al*., 2023; Bi *et al.*, 2023; Lam *et al*., 2022; Chen *et al.*, 2023a; Chen *et al.*, 2023b). Unlike the cautious approach of forecasting only one or two meteorological elements in the first and second stages, these large MWMs produce dozens of meteorological elements globally. The coupling training of these elements may be one of the reasons for the excellent forecasting performance of these large models. In addition, there have been breakthroughs in algorithms, no longer requiring "cubed-sphere mapping to data" or relying solely on CNN for information (features) extraction. Instead, more advanced Transformer or graph neural networks (GNN) models are used to extract information among grid points (especially those that are far apart) and among elements. Consequently, their performance far surpasses the previous two stages of works and can even rival the forecasting performance of the

ECMWF IFS model. Naturally, these large MWMs require more GPU computing resources and consuming large volume of training data.

The fourth stage is the application stage of MWMs. The development of MWMs is still ongoing. In the future, there will undoubtedly be more large MWMs developed. However, one of the most important tasks in the application stage is to test the generalizability of these MWMs and explore their optimal operational usage ways. Therefore, it is necessary to verify the forecasting performance of the MWMs on other commonly used global NWP systems, besides the ECMWF's. For example, using operational analyses of different NWP systems as inputs for the MWMs to assess if the outputs can still be comparable or surpass the performance of corresponding NWP products, and whether the use of high-quality local initial conditions as the model input can improve the accuracy of MWMs. This can help us determine whether MWMs possess generalizability, and promote their practical application.

## 3  Data, Evaluation Metrics, and Testing Environment

**3.1 Data**

The NWP products we used for testing include the ERA5 reanalysis data and the ECMWF-IFS[1] forecasting products from the ECMWF, the NOAA Global Forecast System (GFS) forecasting products generated by the National Weather Service in the United States, the GRAPES forecasting products from the China Meteorological Administration, the YHGSM (Yang *et al*., 2015; 2017) global atmospheric spectral model forecasting products, and the concatenation of two NWP products (to verify whether the improvement of local initial conditions can enhance the forecasting quality).

ERA5 is the fifth-generation ECMWF atmospheric reanalysis product. It includes global atmospheric, land surface and oceanic data since 1940. ERA5 is based on the IFS version Cy41r2, which started operation in 2016, and has been developed

---
[1] The ECMWF-IFS data were tested by Institute of Atmospheric Physics, CAS independently

over several decades to incorporate advancements in atmospheric physics, dynamic mechanisms, and data assimilation. The horizontal resolution of ERA5 is approximately 30 km (0.25° latitude/longitude), and the product outputs are provided on an hourly basis.

ECMWF-IFS is a comprehensive global weather forecasting system. It is widely acknowledged as one of the foremost and most precise forecasting systems currently. The latest version is CY47R3, which consists of 137 vertical model levels, with the model top at 0.01hPa. The horizontal resolution is approximately 13 km (0.125° latitude/longitude). The ECMWF-IFS encompasses a wide range of products, including deterministic and ensemble forecast products. It provides valuable information for meteorological professionals and practitioners.

NOAA-GFS is a weather forecasting system developed by the National Centers for Environmental Prediction (NCEP) of the United States. It utilizes the FV3 (Finite Volume Cubed 3) model with a horizontal resolution of approximately 13 km (0.125° latitude/longitude) and consists of 127 vertical model layers. Through ocean-atmosphere coupling, it offers global atmospheric and oceanic forecast products. NOAA-GFS provides four forecast updates per day (00Z, 06Z, 12Z, and 18Z) and offers hourly forecasts for the first 5 days, followed by three-hourly forecasts from day 5 to day 16.

CMA-GRAPES is a comprehensive NWP system developed by the China Meteorological Administration. It provides a wide range of atmospheric and oceanic meteorological element forecast products, offering global weather forecasts for up to 10 days. The global product has a resolution of approximately 30 km (0.25° latitude/longitude) and consists of 87 vertical model layers, with the model top at 0.1hPa.

YHGSM is a global atmospheric spectral model that has performance parameters similar to ECMWF-IFS. The forecast products we used in this work has a horizontal resolution of 0.5° latitude/longitude and consists of 91 vertical model layers. The products include the forecasts for up to 10 days, with updates produced twice daily.

Since the resolution of input data requirement for the Pangu-Weather model is

0.25°, all aforementioned data are interpolated to 0.25° by using a two-dimensional linear interpolation method. In the experiments, the forecast variables are also consistent with the input variables of the Pangu-Weather model. The surface variables include mean sea level pressure (MSLP), U10 (eastward wind component of the 10m wind), V10 (northward wind component of 10m wind), and T2 (temperature at 2m above ground). The variables at pressure levels include geopotential height (Z), specific humidity (Q), temperature (T), eastward wind component (U), and northward wind component (V). The pressure levels used are 1000, 925, 850, 700, 600, 500, 400, 300, 250, 200, 150, 100, and 50hPa.

**3.2 Evaluation Metrics**

All instances were tested using ERA5 reanalysis as the ground truth, both globally and in the East Asia region (10°S-60°N, 60°E-150°E). Latitude-weighted Root Mean Square Error (RMSE, lower is better) and latitude-weighted Anomaly Correlation Coefficient (ACC, higher is better, Jolliffe and Stephenson 2003) were utilized as evaluation metrics. The ACC measures how accurately the model forecasts departures from climatology.

**3.3 Testing environment**

The testing environment consisted of an NVIDIA Titan RTX GPU (with 24GB VRAM), 48GB RAM, Ubuntu 22.04, CUDA 11.6, and cuDNN 8.2.4. In the experiments, the Pangu-Weather model (Inference by GPU) required approximately 14GB VRAM for single-step forecasting, and the inference time for a single prediction was completed within 4 seconds.

# 4 Results

In this study, three instances were selected to test the Pangu-Weather model: the forecast products were initialized at 00:00UTC on June-6, 2023, June-16, 2023, and June-26, 2023. The operational analyses at the initial time (00:00UTC) of different NWP products are used as the inputs for the Pangu-Weather model. The Pangu-Weather model is run for 10 days (240 hours) forecasts for all these instances. The results were compared at 24-hour intervals. We select the results for the first instance to conduct detailed analysis. More results of these instances are listed in the appendix.

**4.1 Visual display**

Firstly, to visually demonstrate the compatibility of the Pangu-Weather model to operational data, Figure 1 presents the examples of the forecasts by the Pangu-Weather model for the 72 hour lead time from the initial time on June 6, 2023. It includes four surface variables: MSLP (unit: Pa), T2 (unit: K), U10 (unit: m/s), V10 (m/s) and five 500hPa variables: Q500 (unit: g/kg), T500 (unit: K), U500 (unit: m/s), V500 (unit: m/s), Z500 (unit: m$^2$/s$^2$).

As shown in Figure 1, it is evident that when using ERA5 reanalysis and GFS operational analysis as the input, the Pangu-Weather model generated the forecasts similar to the ERA5 data as the reference. The visual differences between the two products are minimal.

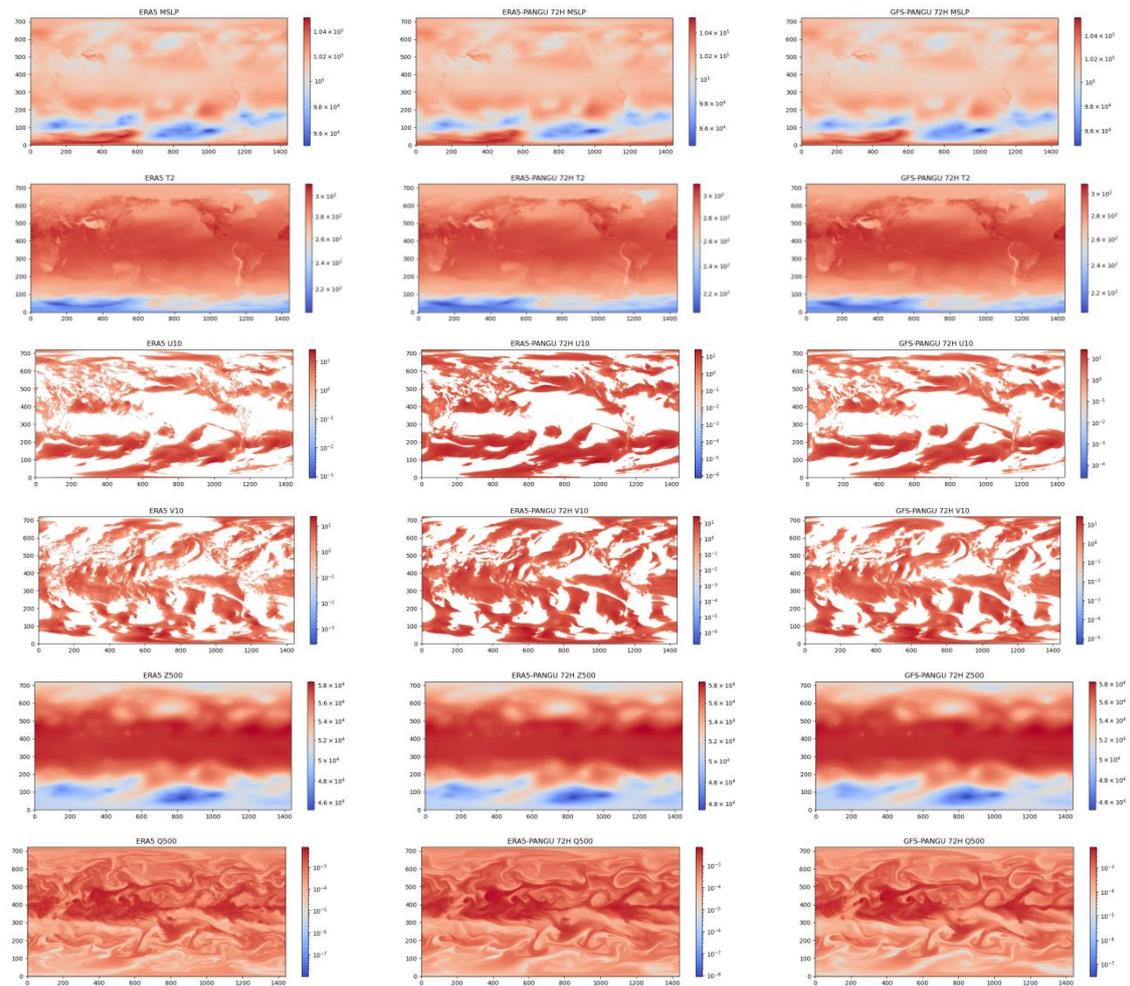

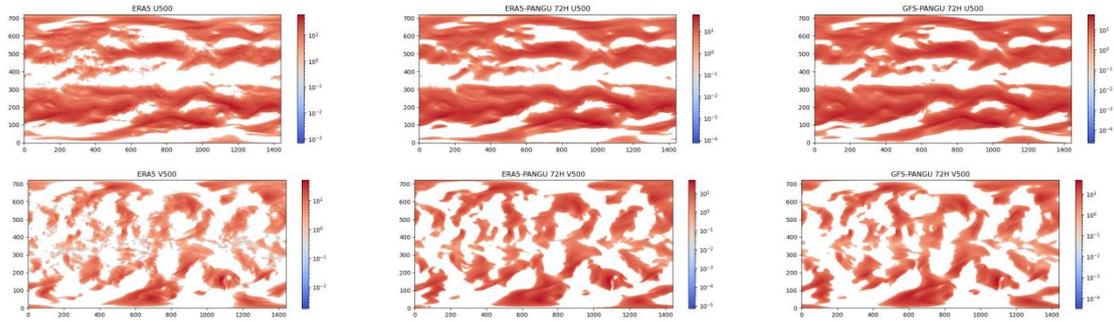

Figure 1. The first column indicates the ground truth of the ERA5 reanalysis data for the forecast time (00h UTC on June 9). The second column and third column indicate the forecasts for the 72 hour lead time (00h UTC on June 9) using the ERA5 reanalysis product and the NOAA-GFS operational analysis as the initial conditions of the Pangu-Weather model, respectively.

**4.2 Comparison of results**

(1) Comparison of forecasts between NWP systems and Pangu-Weather model initialized from the same initial conditions.

Figures 2 to 10 depict the comparison between the forecasts of the NWP model and the Pangu-Weather model using the same initial conditions at 00:00 UTC on June 6, 2023. These figures display the latitude-weighted root mean square error (RMSE) and latitude-weighted anomaly correlation coefficient (ACC) comparison for nine variables: MSLP, T2, U10, V10, Q500, T500, U500, V500, Z500. To ensure fair comparisons, the Pangu-Weather model was compared with GFS, GSM, and GRAPES models using their own initial conditions.

For the first column, it shows the results from the GFS products (gfs-nwp) and the Pangu-Weather model (gfs-pangu) using the GFS operational analysis as the initial conditions. For the second column, it shows the results from the YHGSM products (GSM-nwp) and the Pangu-Weather model (GSM-pangu) using the YHGSM operational analysis as the initial conditions. For the third column, it shows the results from the GRAPES products (GRAPES-nwp) and the Pangu-Weather model (GRAPES-pangu) using the GRAPES operational analysis as the initial conditions.

The first row represents the latitude-weighted RMSE (Root Mean Square Error) metric, and the second row represents the latitude-weighted ACC (Anomaly

Correlation Coefficient) metric.

From these figures, it can be observed that:

1) When using the operational analyses of various NWP products as input data for the Pangu-Weather model, the overall quality of the Pangu-Weather model's products is generally comparable or superior to the corresponding NWP products. This indicates that the Pangu-Weather model can be used to generate products for practicable operational forecasting. This addresses the first question in the introduction, which is whether the operational analyses of different NWP systems can be applicable to the large MWMs and achieve forecasting performance comparable to, or even surpassing, the corresponding NWP products.

The results of the Pangu-Weather model show significant improvement compared to the GFS and YHGSM products. However, compared to the GRAPES products, the results of the Pangu-Weather model demonstrate both improvement and degradation. We speculate that this phenomenon may be due to the data-driven model based on ML being trained to fit the skills of the ECMWF NWP systems (i.e. ECMWF-IFS version Cy41r2, 2016), which used to generate the training dataset. Therefore, the MWM is better able to adapt to the initial conditions generated by specific NWP systems, resulting in quality differences of the forecasting products compared to the corresponding NWP products.

Furthermore, when using ERA5 reanalysis products as initial conditions for the Pangu-Weather model, the forecasts of the Pangu-Weather model generally outperform the forecasts obtained by using the operational analyses of YHGSM, GFS, and GRAPES as input data for the Pangu-Weather model. This indicates that the quality of the initial conditions of the Pangu-Weather model determines the performance. Improving the overall quality of the initial conditions contributes to enhancing the forecasting performance of the large MWM.

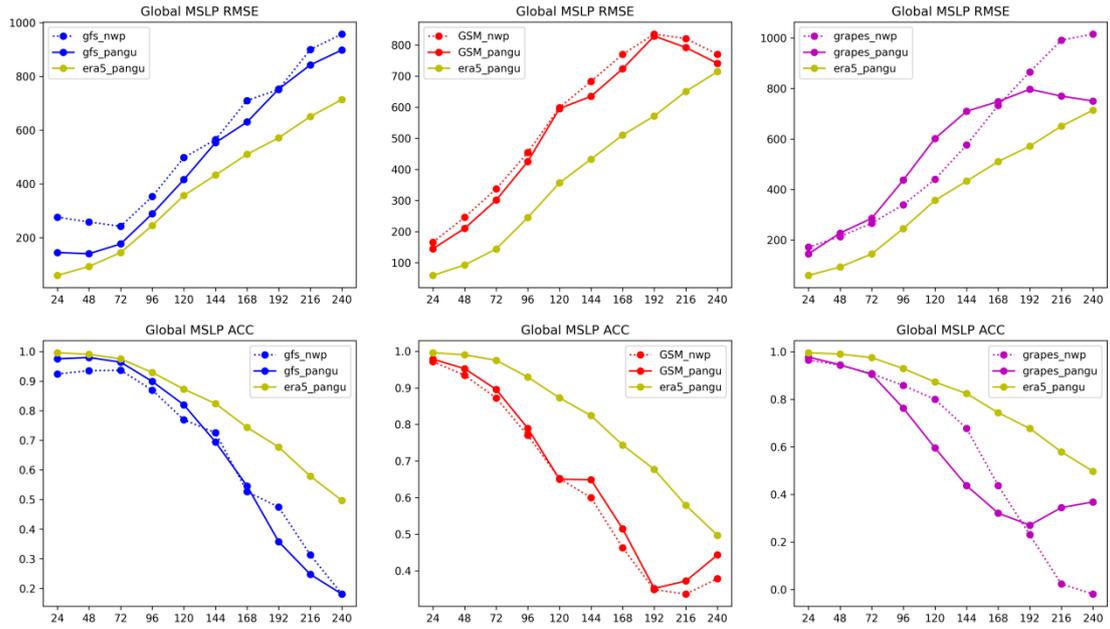

Figure 2. Forecasting performance for MSLP of NWP systems and Pangu-Weather model with the same initial conditions.

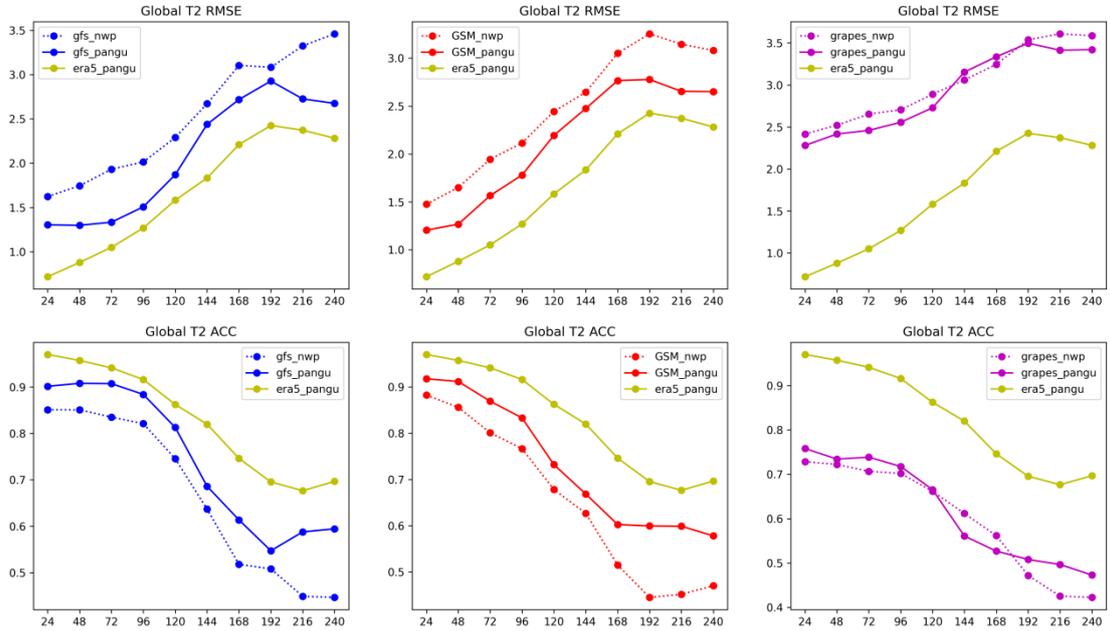

Figure 3. Same as Figure 2, but for T2.

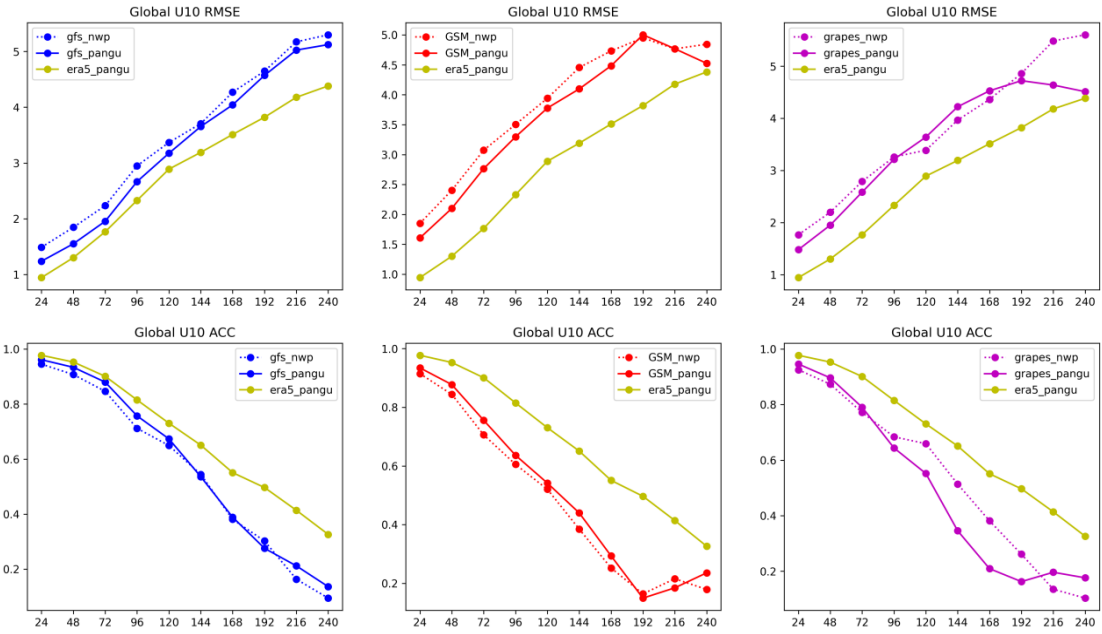

Figure 4. Same as Figure 2, but for U10.

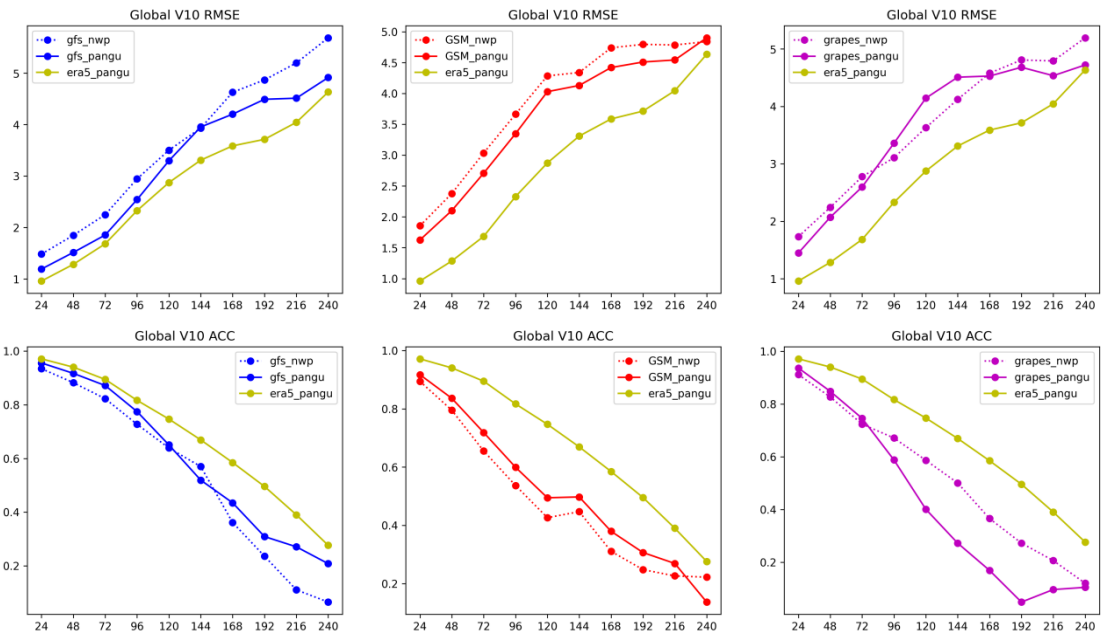

Figure 5. Same as Figure 2, but for V10.

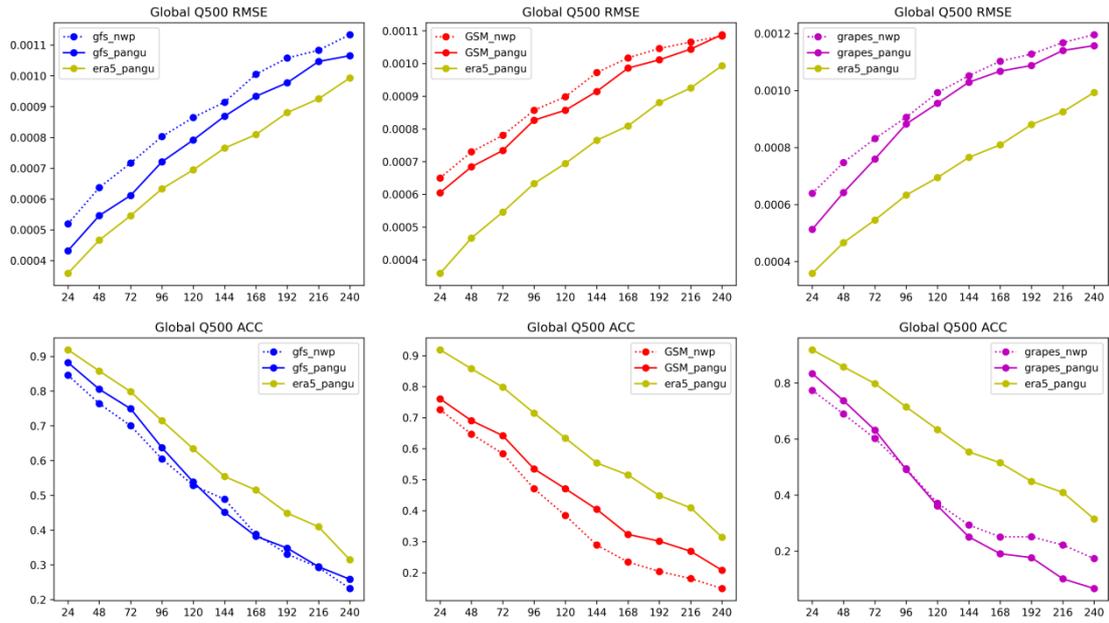

Figure 6. Same as Figure 2, but for Q500.

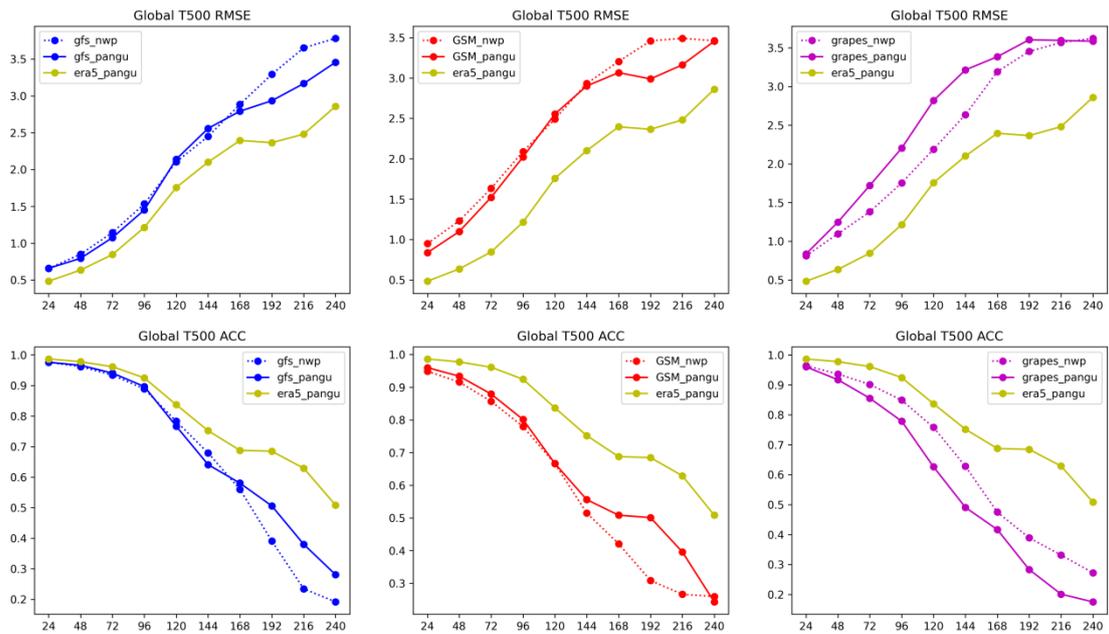

Figure 7. Same as Figure 2, but for T500.

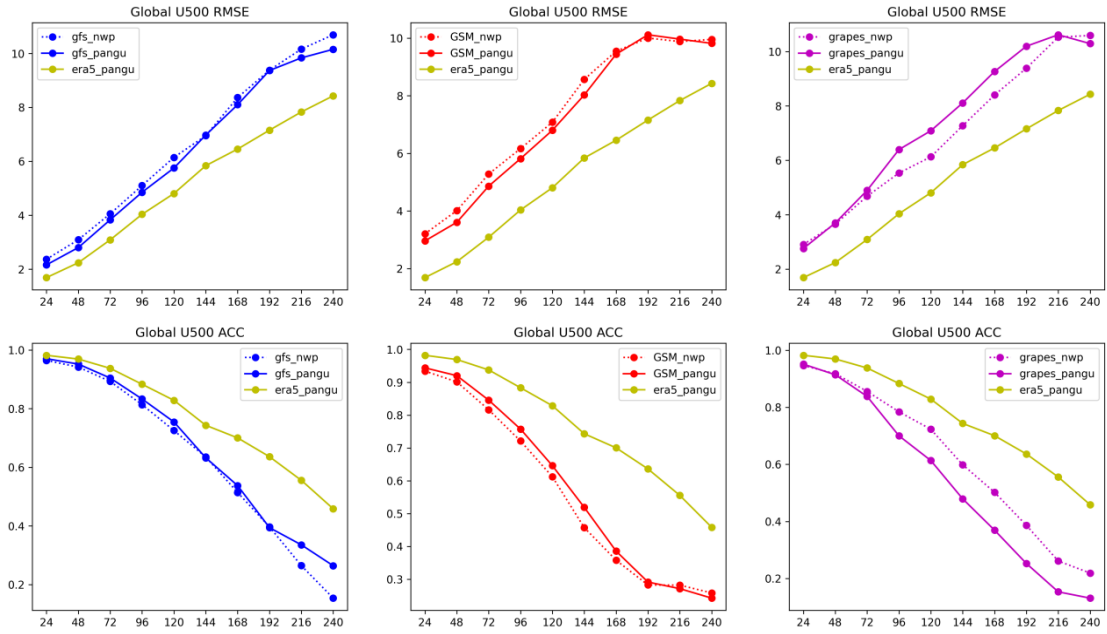

Figure 8. Same as Figure 2, but for U500.

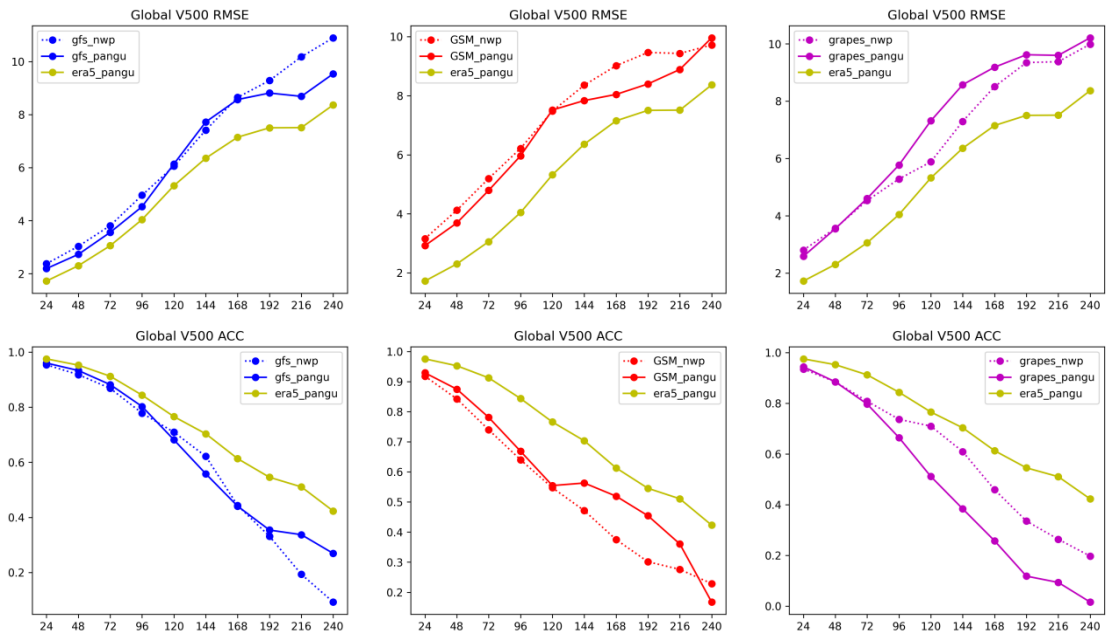

Figure 9. Same as Figure 2, but for V500.

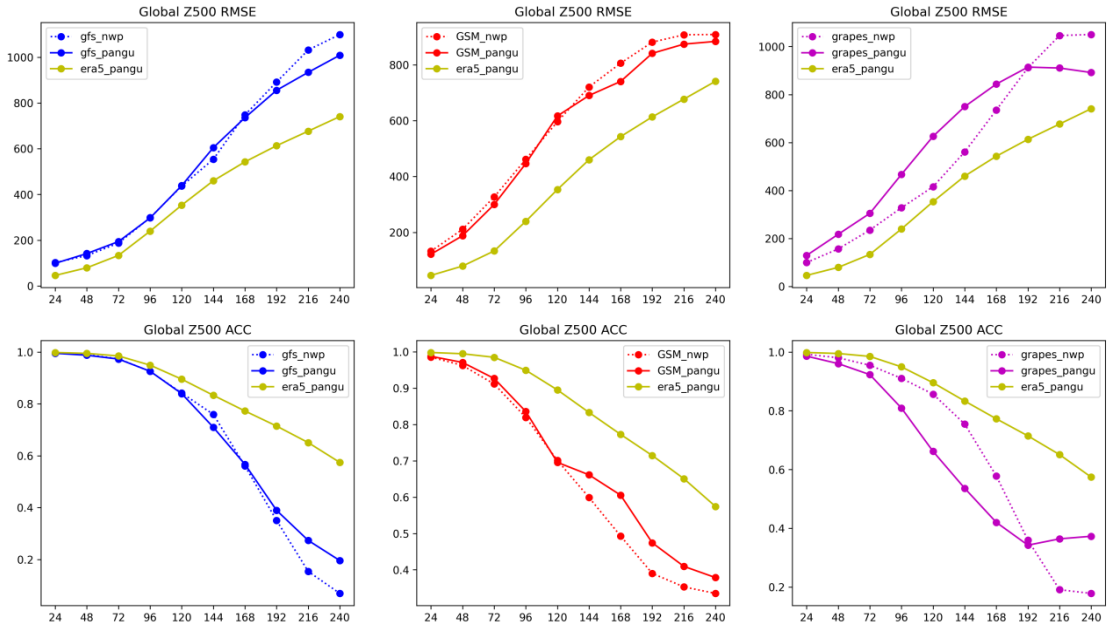

Figure 10. Same as Figure 2, but for Z500.

(2) The impact of improving the quality of local initial conditions on the forecasting performance of the Pangu-Weather model.

This study also verified that improving the quality of local initial conditions can enhance the forecasting performance of the Pangu-Weather model.

As depicted in Figures 11 to 19, the first column depicts the comparison between the ECMWF-IFS and NOAA-GFS for the East Asia region. Results show that the forecasts from ECMWF-IFS outperform those from NOAA-GFS in this instance. Building upon this observation, we merged the pressure level operational analyses of ECMWF-IFS and NOAA-GFS. Specifically, ECMWF-IFS provides the initial conditions (pressure levels) for the East Asia region (10°S-60°N, 60°E-150°E), while NOAA-GFS offers the initial conditions for other regions. By amalgamating these two datasets, a new combined global data of initial conditions, "ecmfpadGFS", was obtained. It can be considered as an improved version of the initial conditions data that deliver better quality for the local (East Asia) region compared to NOAA-GFS.

The second column presents the comparison of global forecasts generated by using the GFS original initial conditions and the concatenated data initial conditions (i.e. ecmfpadGFS) as inputs for the Pangu-Weather model. It was found that improving the quality of local initial conditions (in the East Asia region) can slightly enhance the quality of global forecasting performance generated by the Pangu-Weather model.

The third column represents the comparison of forecasts for the East Asia region generated by using the GFS original initial conditions and the concatenated initial conditions as inputs for the Pangu-Weather model. It can be observed that improving the quality of local initial conditions (specifically in the East Asia region) can significantly enhance the forecasting performance of the Pangu-Weather model for that specific region (East Asia).

This addresses the second question raised in our introduction: for the Pangu-Weather model, achieving better results can be accomplished by improving the quality of local initial conditions which are used as inputs. Therefore, if we aim to enhance the performance of local forecasts for the Pangu-Weather model, it is easier to achieve by improving the quality of local initial conditions (such as through assimilating local observations) rather than improving the quality of global initial conditions.

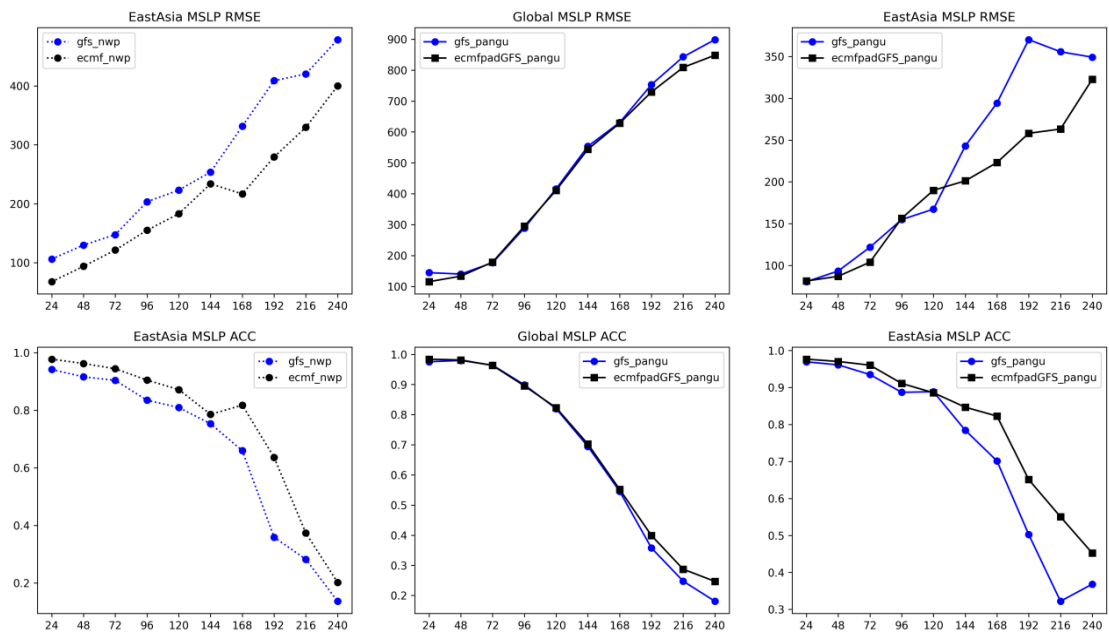

Figure 11. Forecasting performance for MSLP of NWP systems and Pangu-Weather model for East Asia region (left and right) and global (middle).

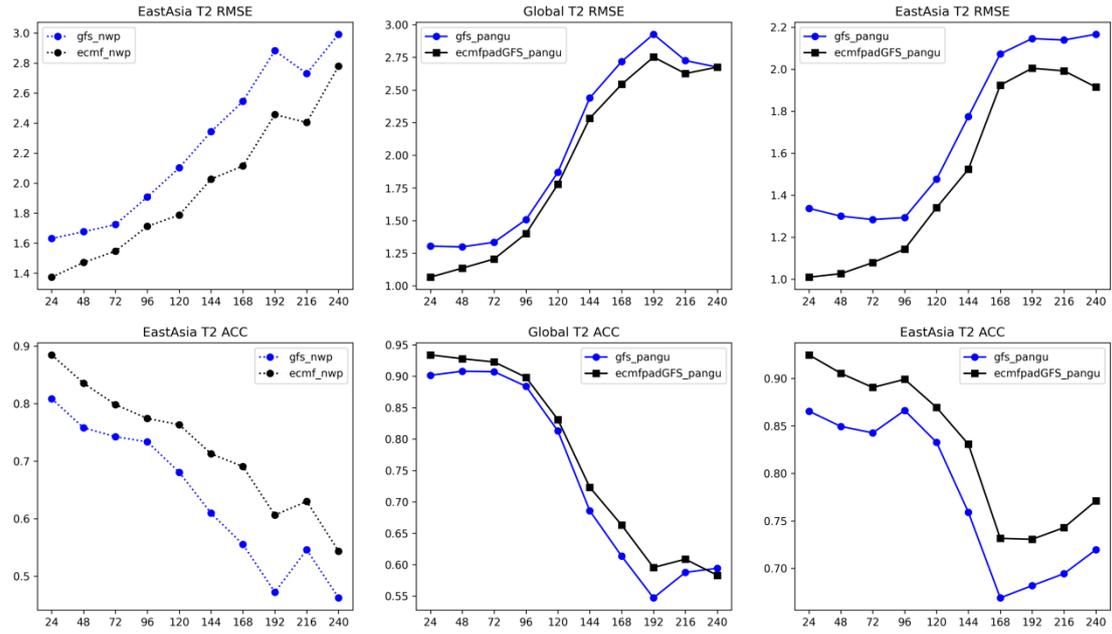

Figure 12. Same as Figure 11, but for T2.

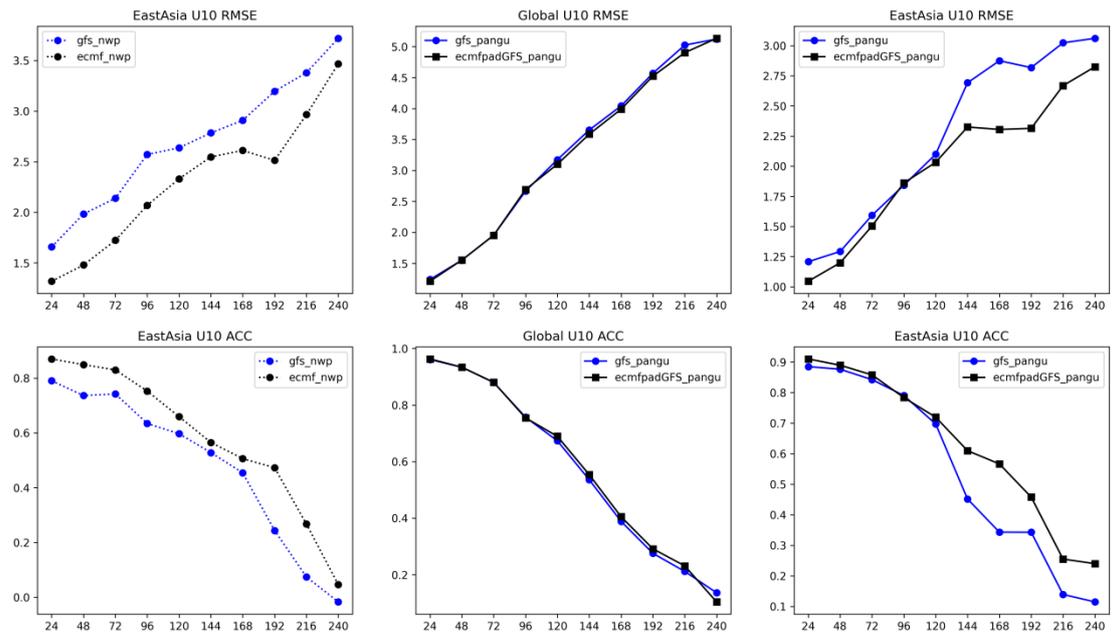

Figure 13. Same as Figure 11, but for U10.

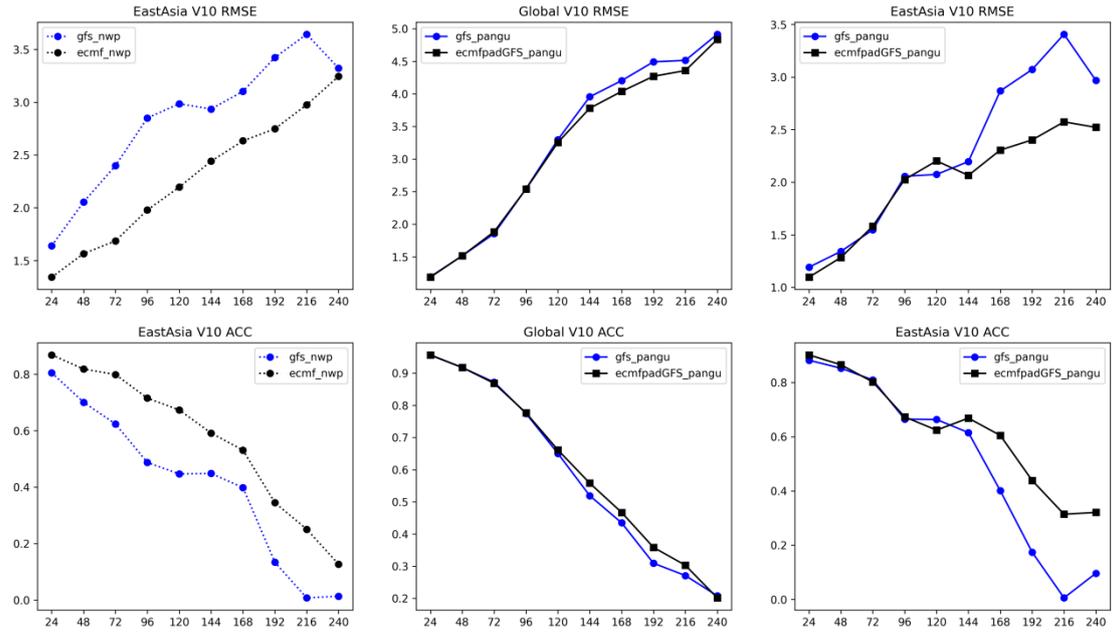

Figure 14. Same as Figure 11, but for V10.

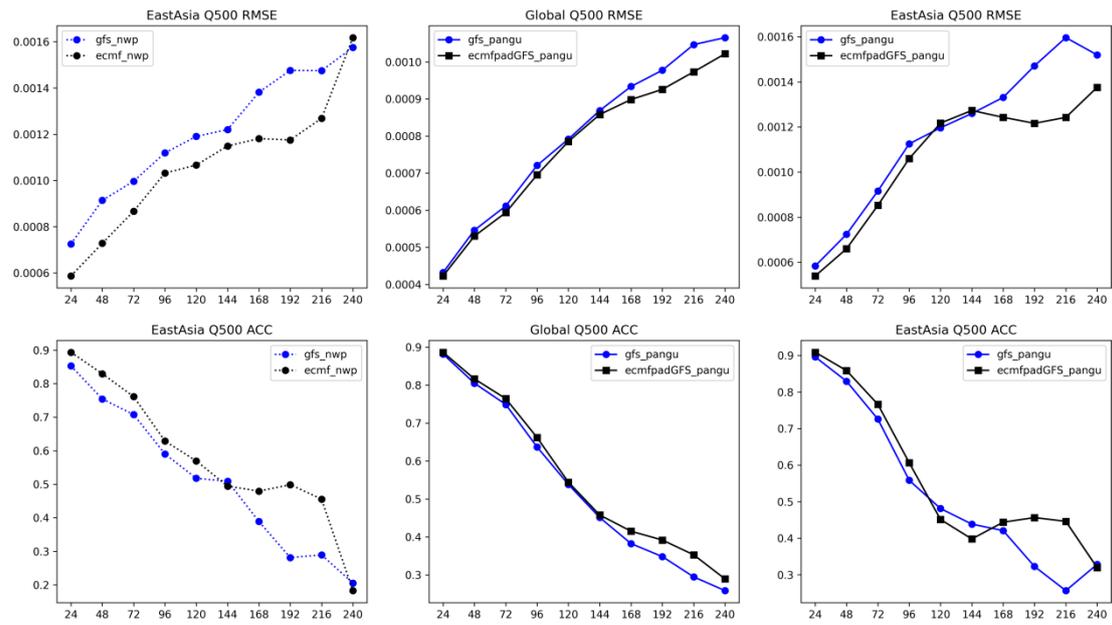

Figure 15. Same as Figure 11, but for Q500.

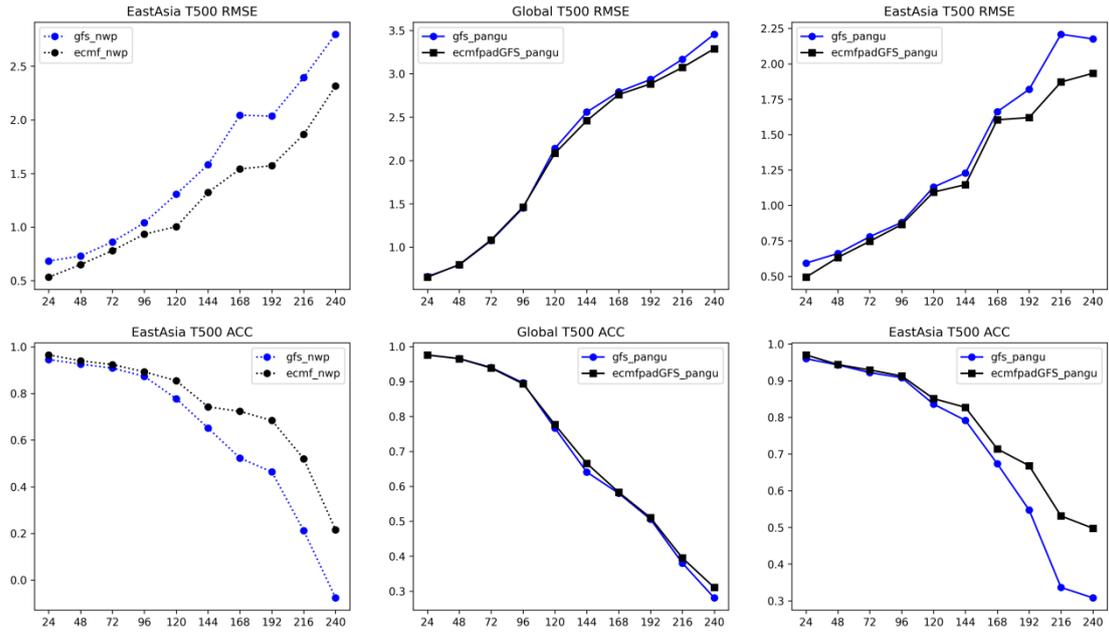

Figure 16. Same as Figure 11, but for T500.

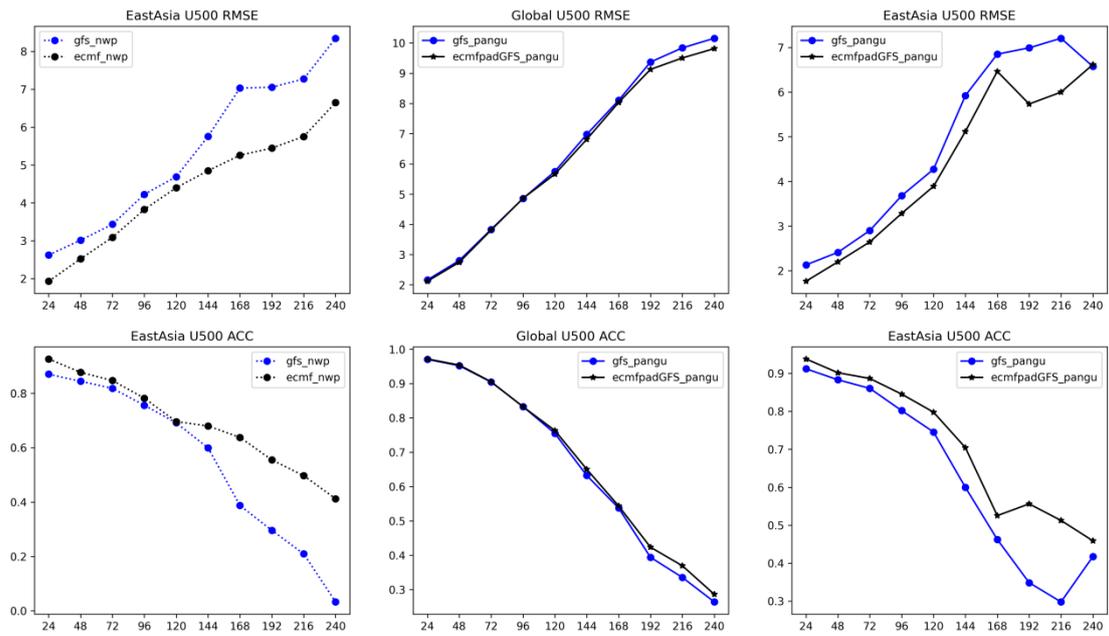

Figure 17. Same as Figure 11, but for U500.

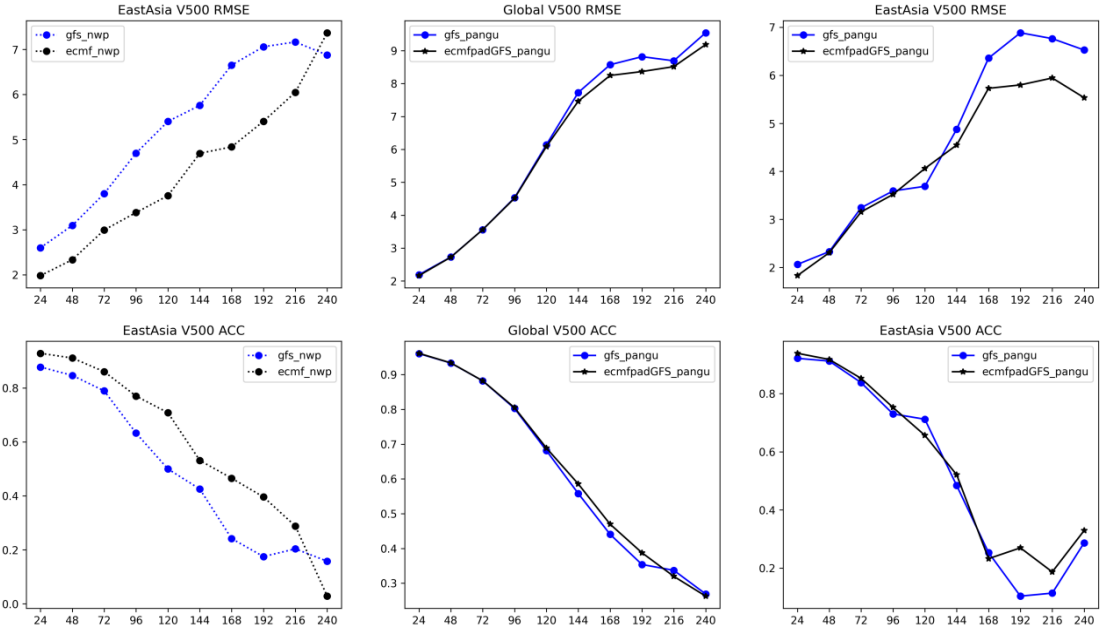

Figure 18. Same as Figure 11, but for V500.

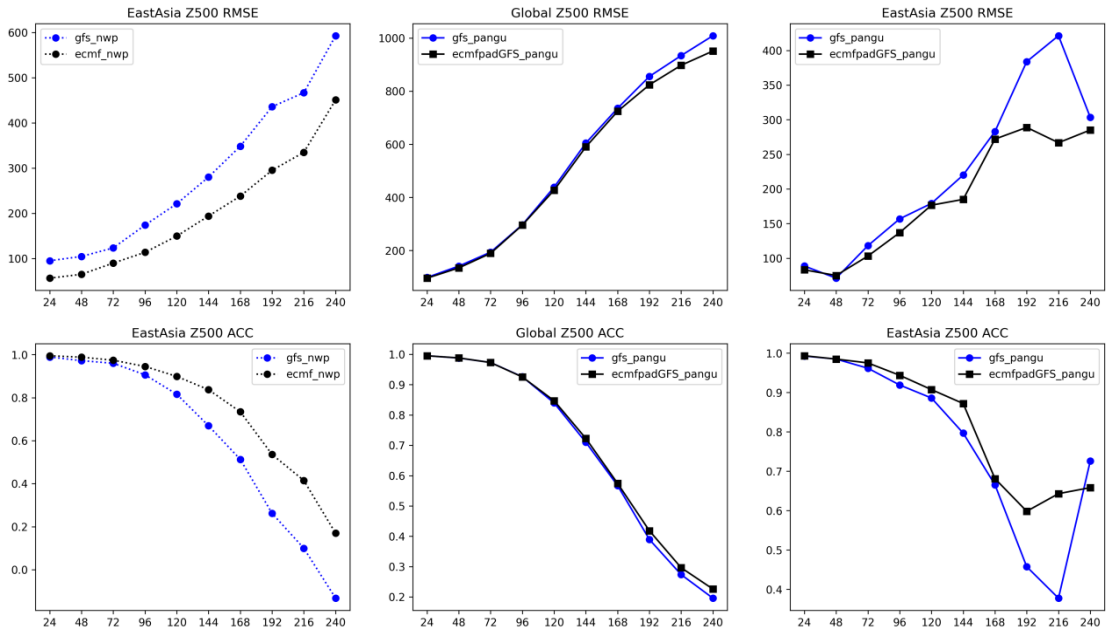

Figure 19. Same as Figure 11, but for Z500.

## 5 Conclusion and Future Work

This paper analyzes the compatibility between the Pangu-Weather model and different meteorological NWP operational data. The results obtained from the case study indicate that:

(1) When using the operational analyses from various operational NWP systems

as inputs, the Pangu-Weather model can generate results that are comparable or superior to the original NWP products.

(2) Improving the overall quality of the initial conditions helps improve the forecasting performance of the Pangu-Weather model.

(3) When using the same initial conditions, Pangu-Weather model's forecasts are better than the NWP system's forecasts in some cases, while in others, they are comparable to the NWP system's forecasts. We speculate that this phenomenon may be due to the Pangu-Weather model being trained to fit the skills of the ECMWF NWP systems (i.e. ECMWF-IFS version Cy41r2, 2016), which used to generate the training dataset, Therefore, the model is better able to adapt to the initial conditions generated by specific NWP systems.

(4) Optimizing the initial conditions in local region (e.g. East Asia region) can enhance the forecasting performance of the Pangu-Weather model for that specific region.

Pangu-Weather model and other data-driven models offer a new approach to weather forecasting. Compared to traditional NWP systems, they have advantages such as lightweight deployment and fast computation. This may lead to various new applications of using meteorological forecasts. Testing their compatibility with operational meteorological data is an important step for understanding their capabilities and conducting further research. This experiment analysis was conducted on a limited number of case data by now. In the future, we will analyze the performance of data-driven forecast models over longer time periods, such as months, seasons, years, or even longer periods. Additionally, as other data-driven weather forecast models do not currently share the inference interfaces, we will further follow up on testing experiments if such interfaces become available in the future.

# Appendix

## 1. Results of 2023.06.06

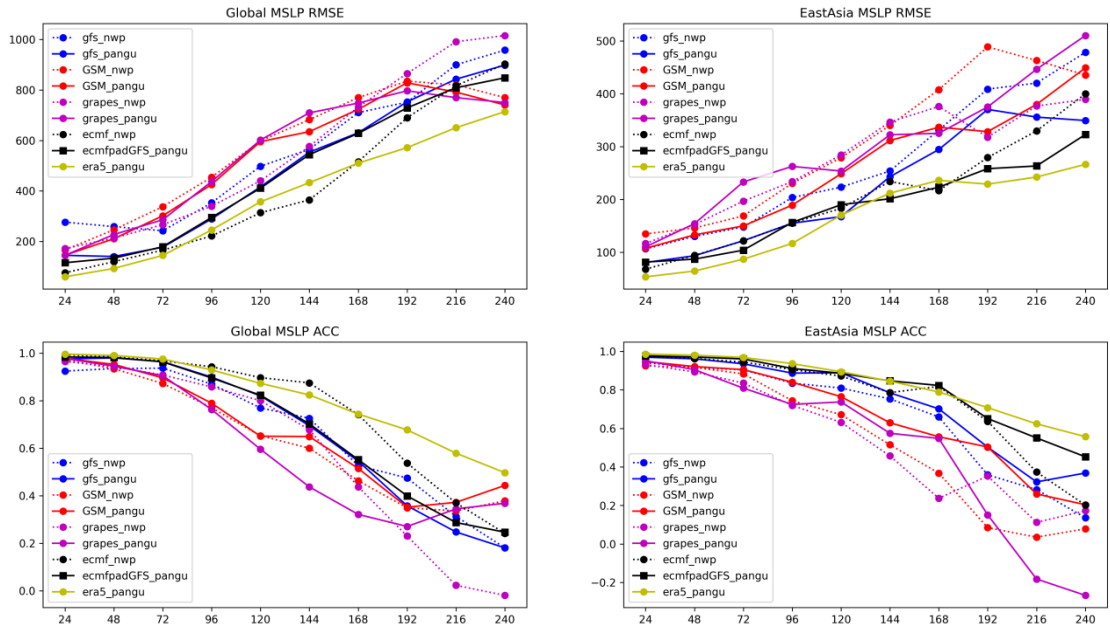

Figure A1.1 Forecasting performance for MSLP of each NWP system and Pangu-Weather model with the same initial conditions for East Asia (right) and globally (left). The forecasts are from the initial time on June 6, 2023.

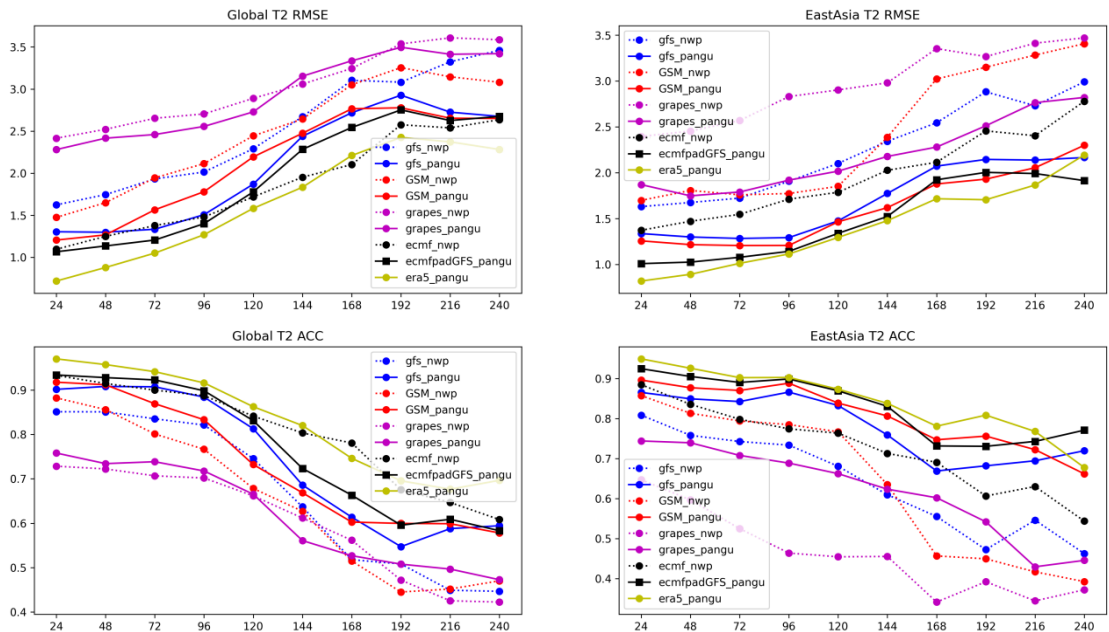

Figure A1.2 Same as Figure A1.1, but for T2.

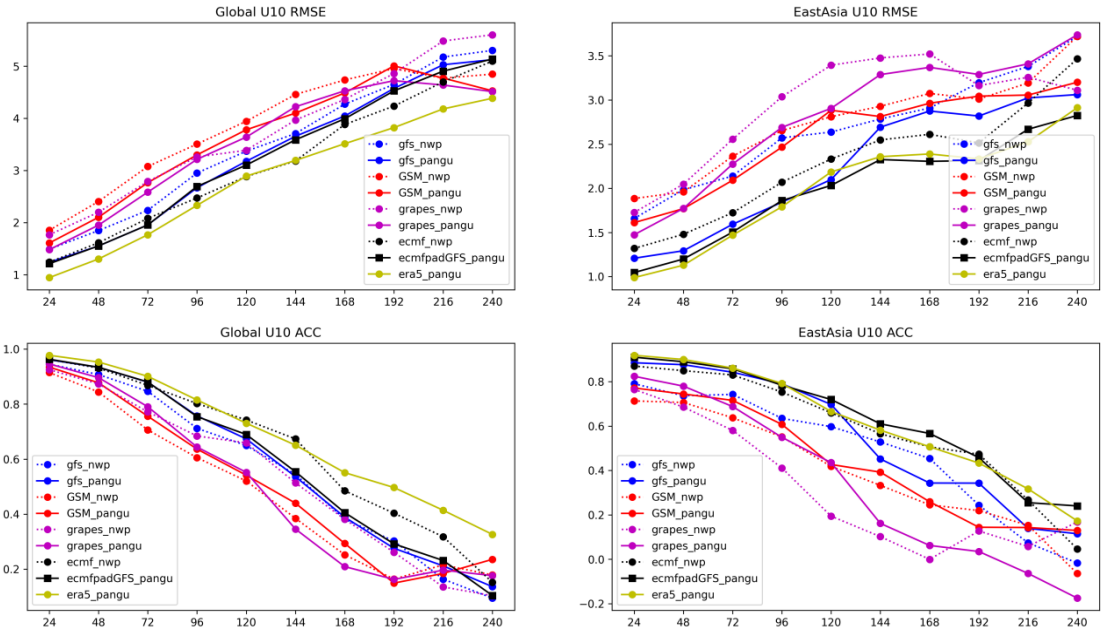

Figure A1.3 Same as Figure A1.1, but for U10.

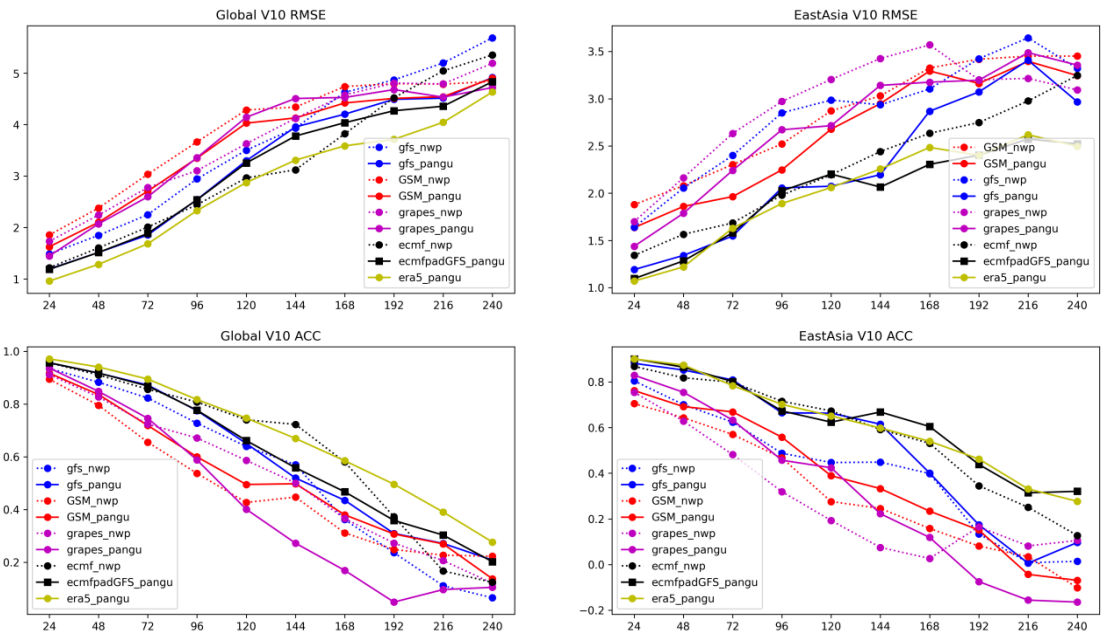

Figure A1.4 Same as Figure A1.1, but for V10.

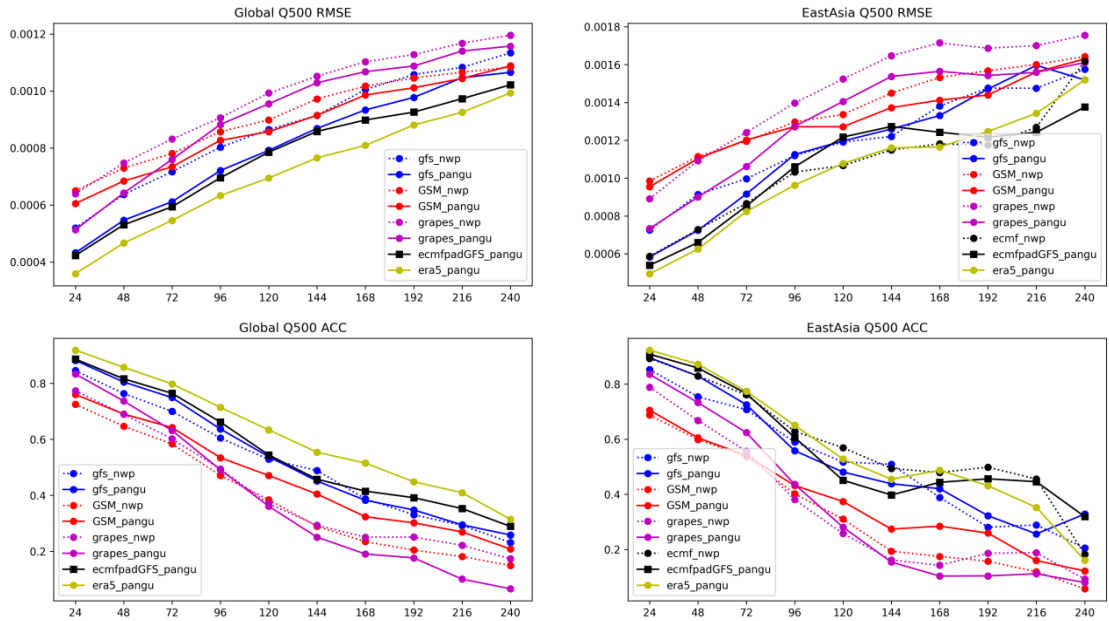

Figure A1.5 Same as Figure A1.1, but for Q500.

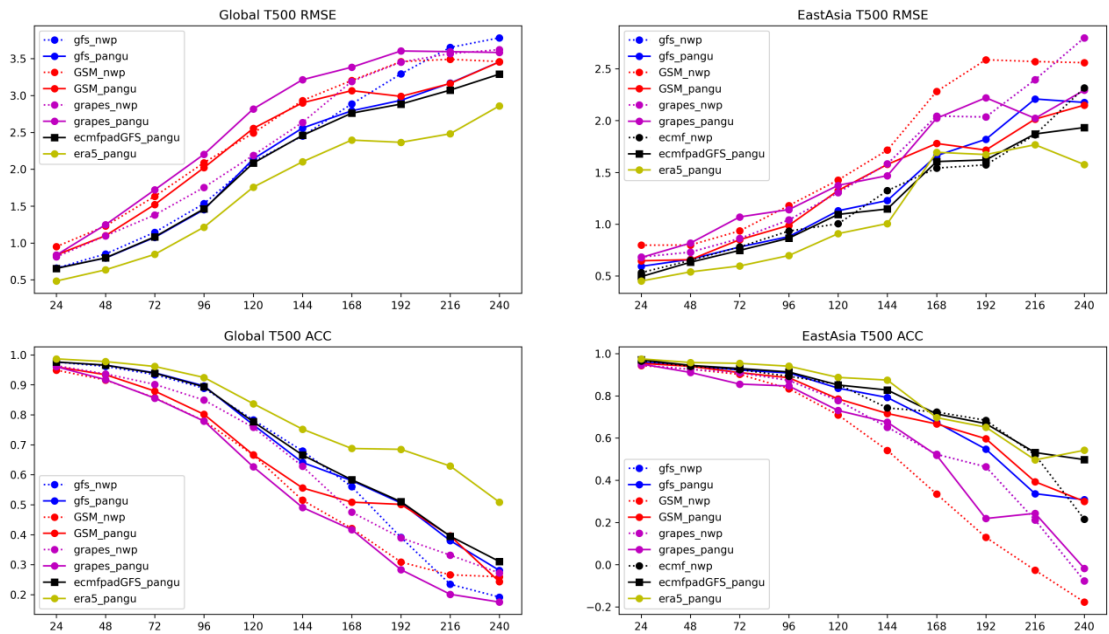

Figure A1.6 Same as Figure A1.1, but for T500.

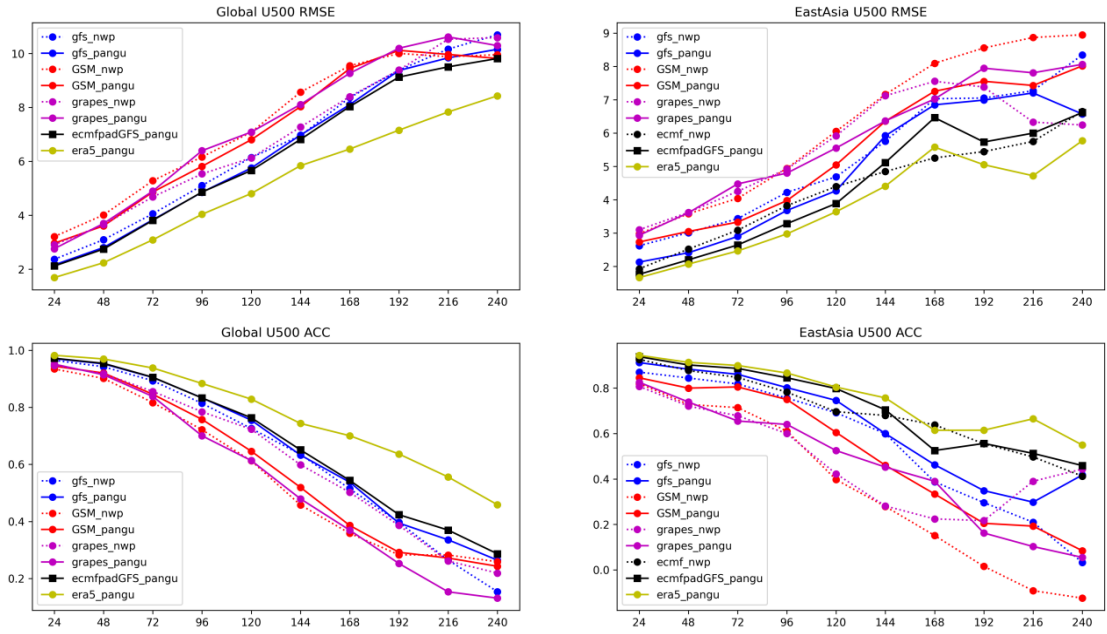

Figure A1.7 Same as Figure A1.1, but for U500.

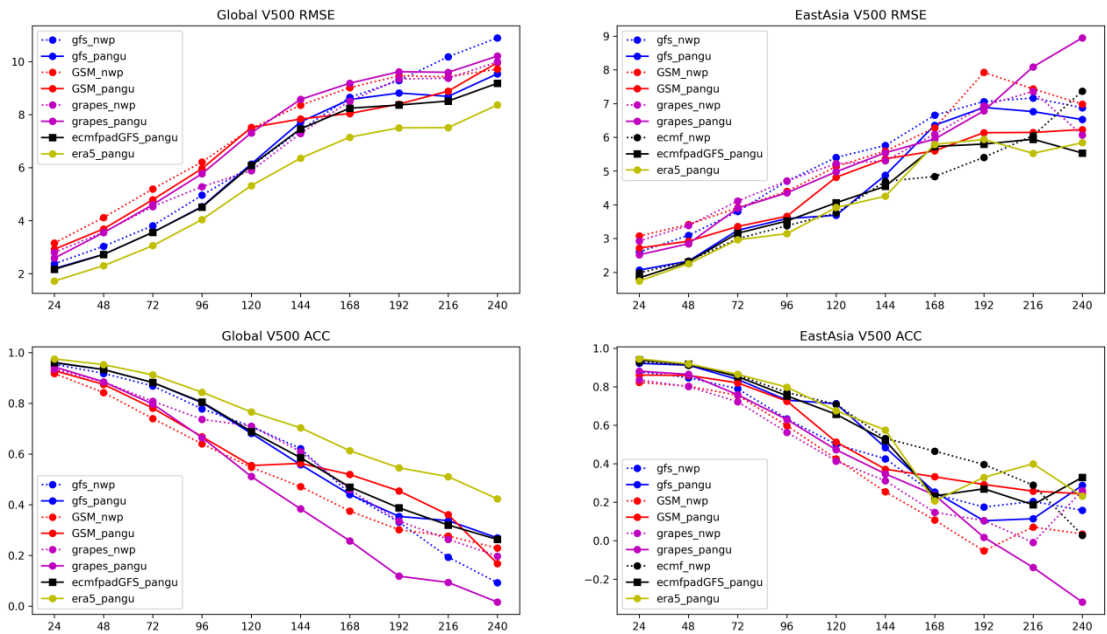

Figure A1.8 Same as Figure A1.1, but for V500.

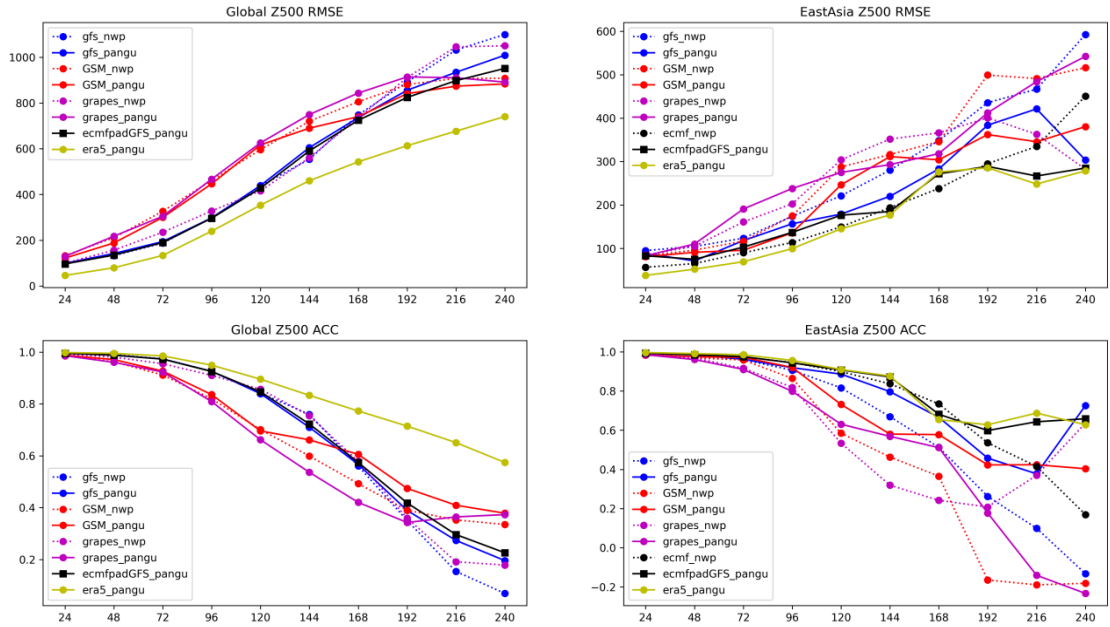

Figure A1.9 Same as Figure A1.1, but for Z500.

## 2. Results of 2023.06.16

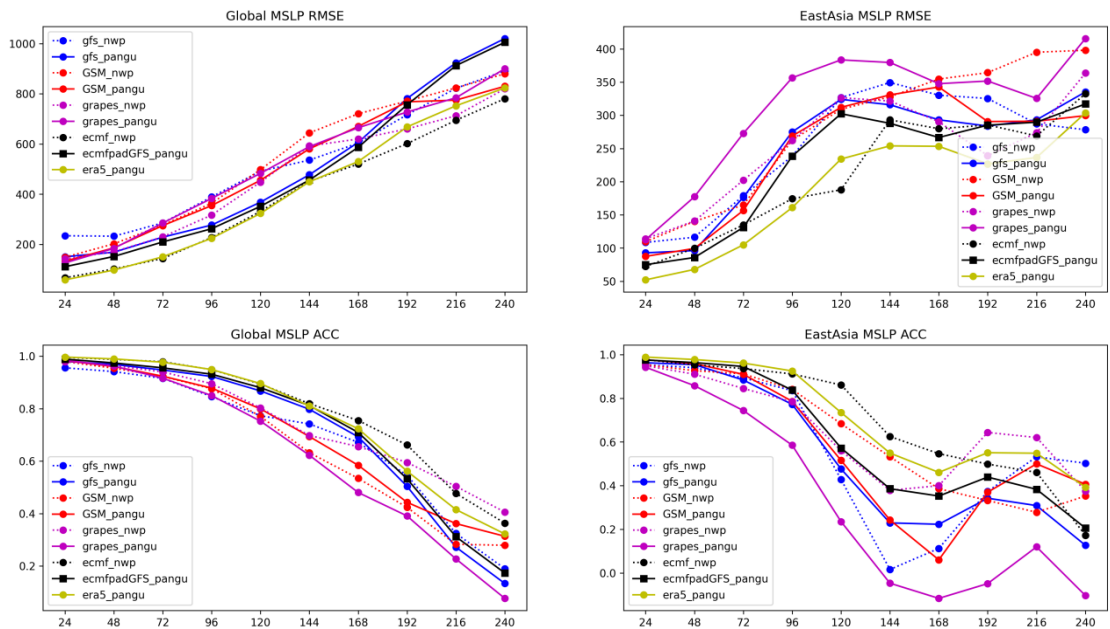

Figure A2.1 Forecasting performance for MSLP of each NWP system and Pangu-Weather model with the same initial conditions for East Asia (right) and globally (left). The forecasts are from the initial time on June 16, 2023.

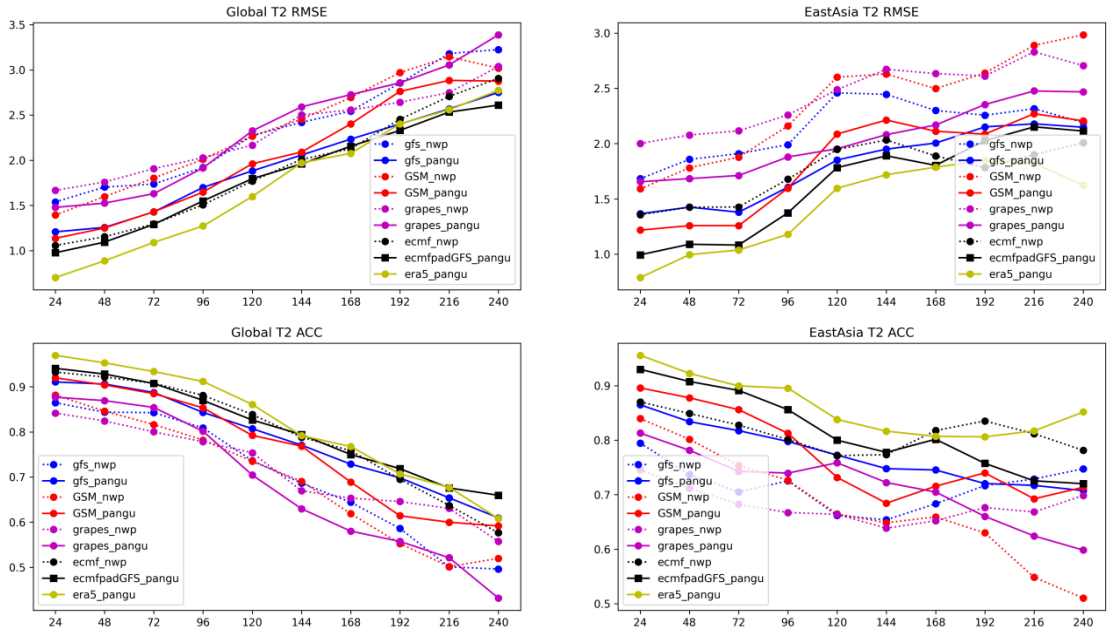

Figure A2.2 Same as Figure A2.1, but for T2.

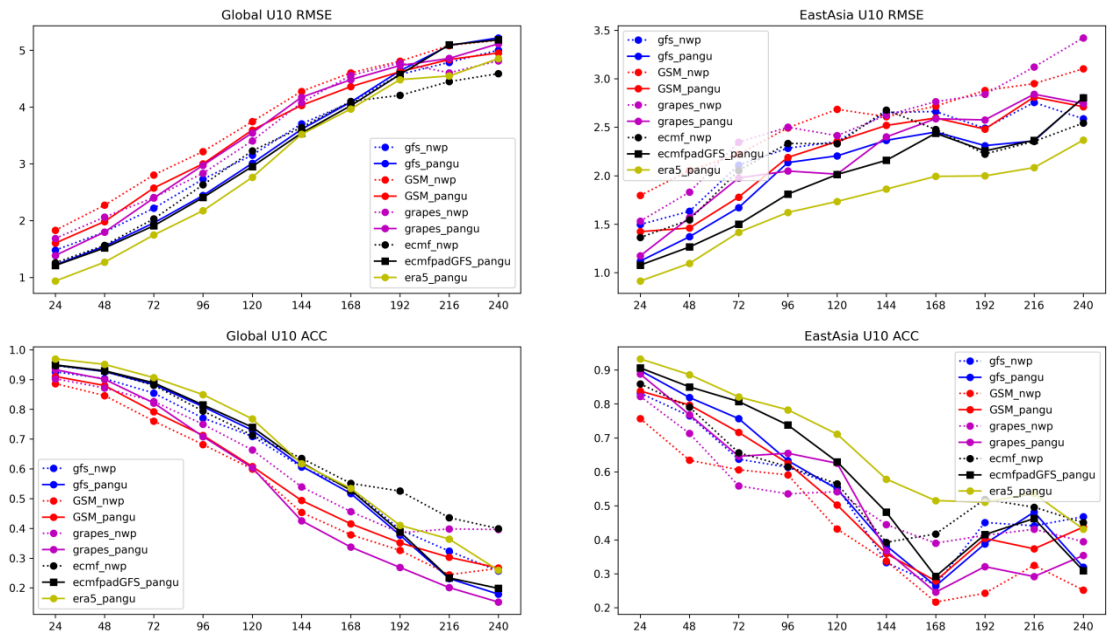

Figure A2.3 Same as Figure A2.1, but for U10.

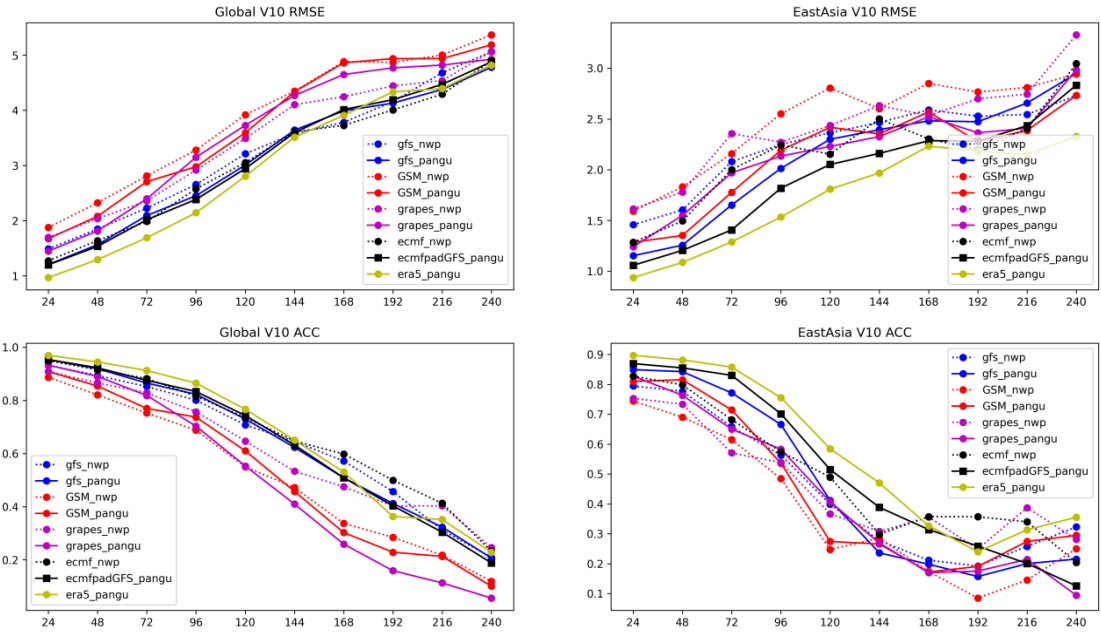

Figure A2.4 Same as Figure A2.1, but for V10.

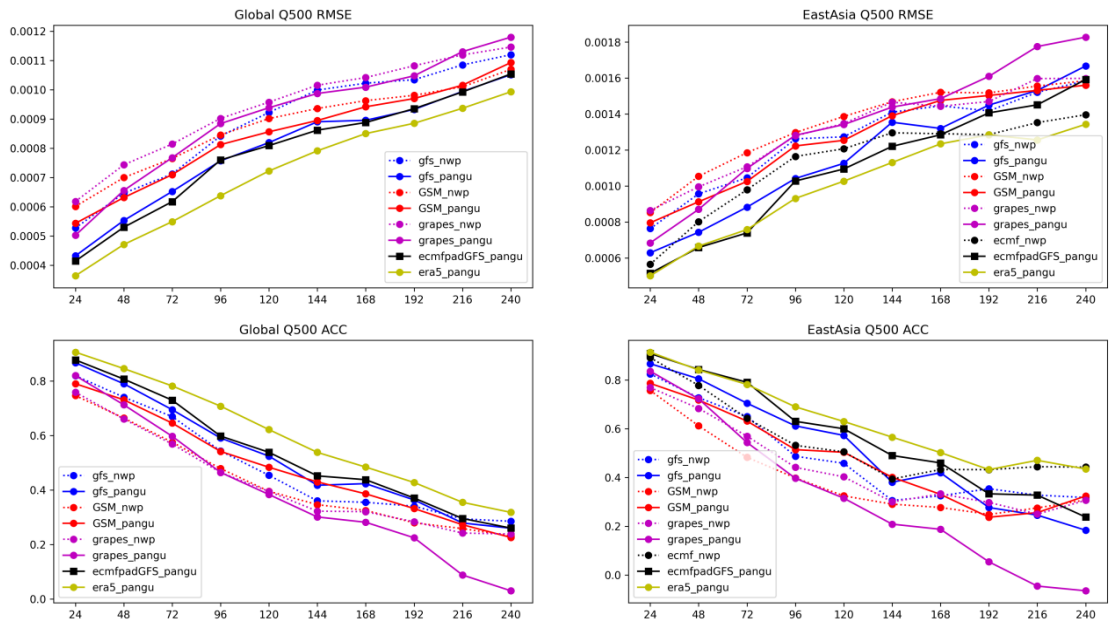

Figure A2.5 Same as Figure A2.1, but for Q500.

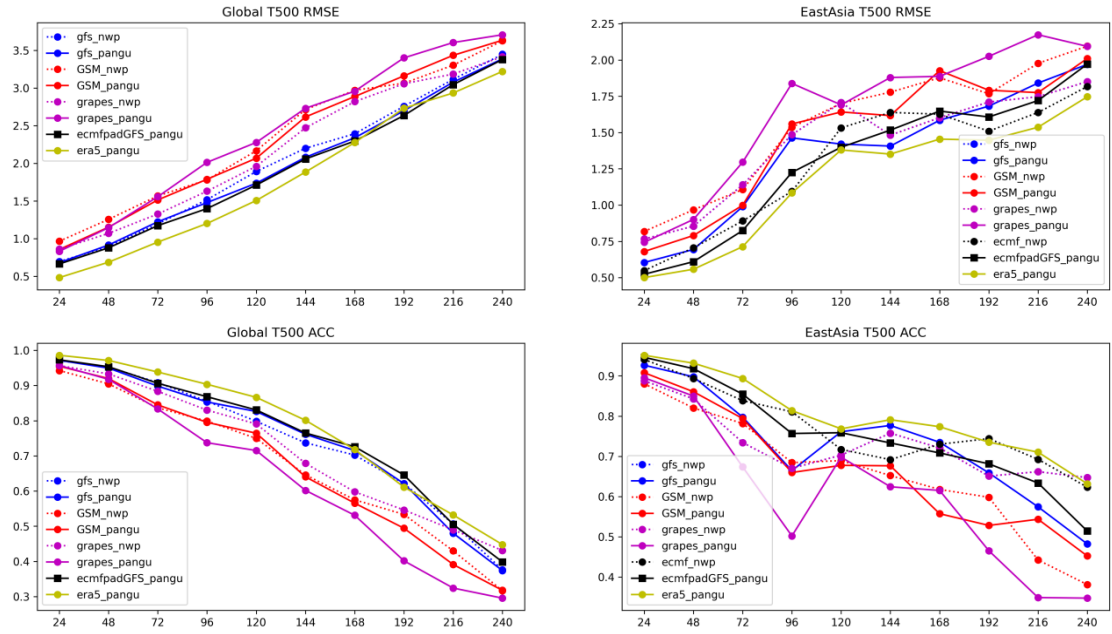

Figure A2.6 Same as Figure A2.1, but for T500.

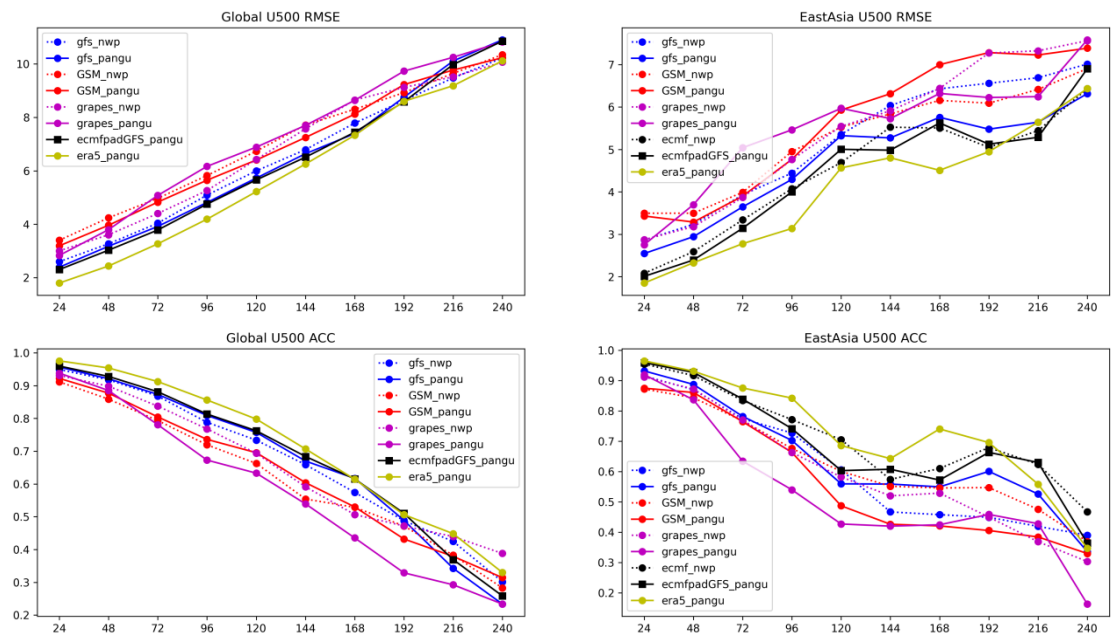

Figure A2.7 Same as Figure A2.1, but for U500.

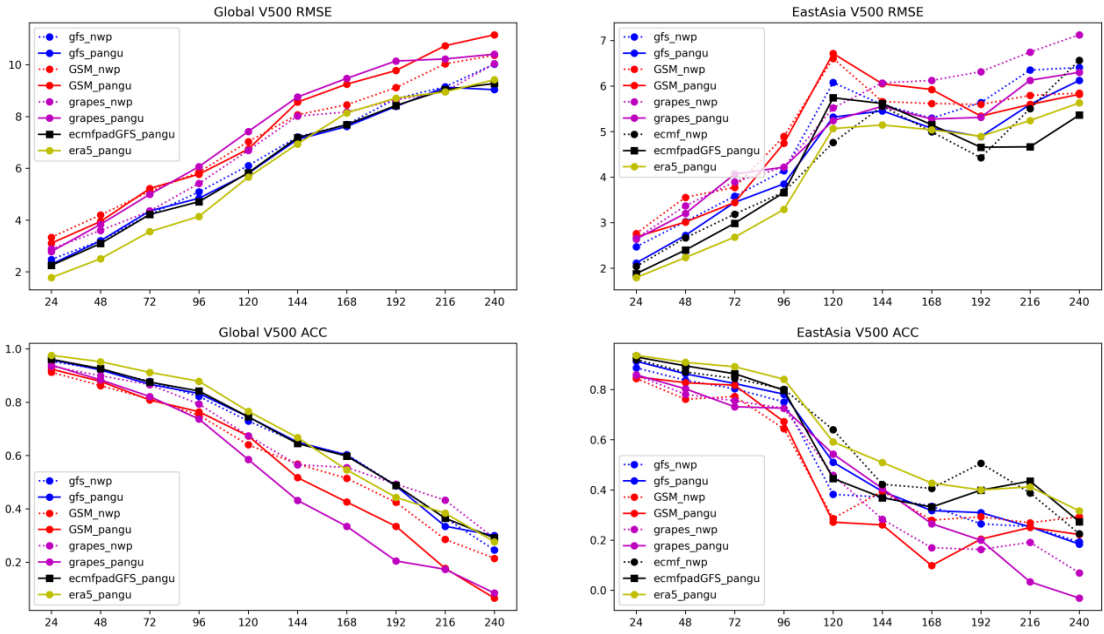

Figure A2.8 Same as Figure A2.1, but for V500.

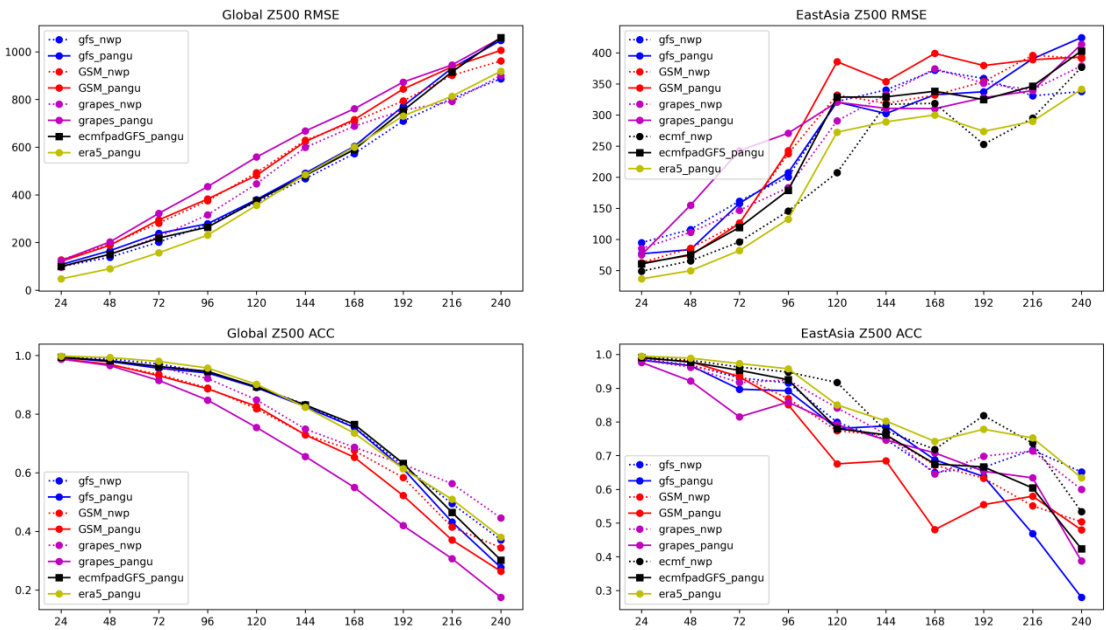

Figure A2.9 Same as Figure A2.1, but for Z500.

## 3. Results of 2023.06.26

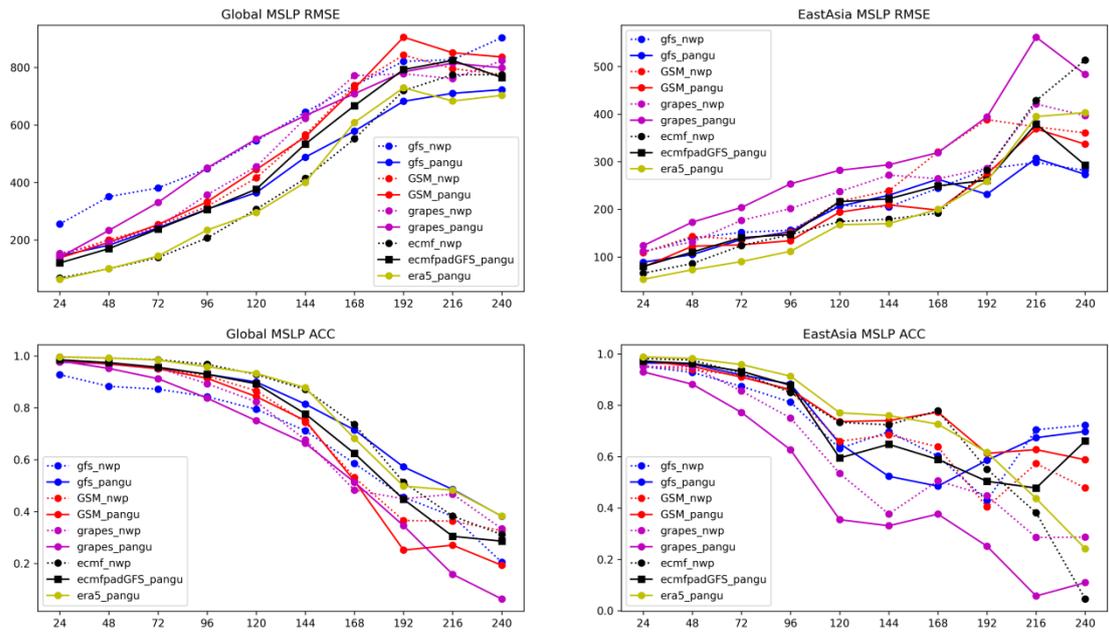

Figure A3.1 Forecasting performance for MSLP of each NWP system and Pangu-Weather model with the same initial conditions for East Asia (right) and globally (left). The forecasts are from the initial time on June 26, 2023.

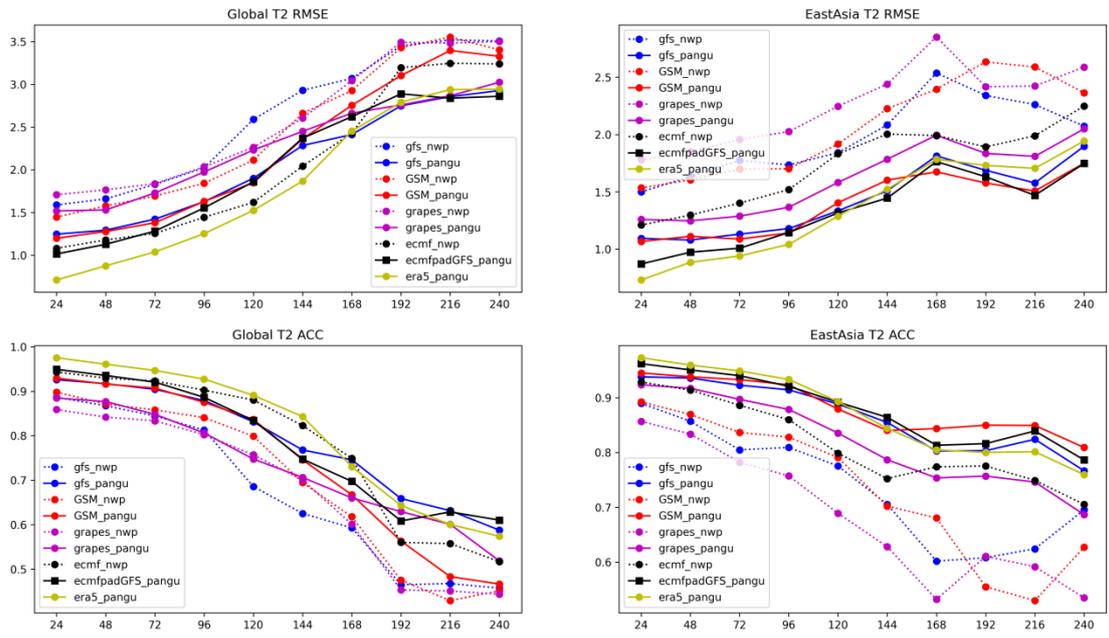

Figure A3.2 Same as Figure A3.1, but for T2.

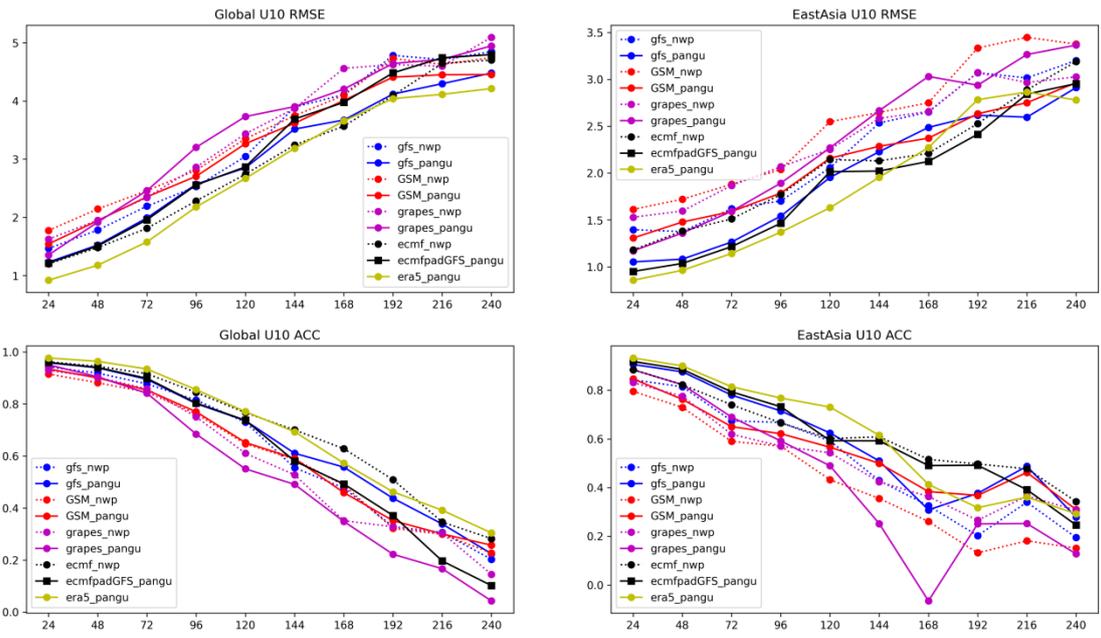

Figure A3.3 Same as Figure A3.1, but for U10.

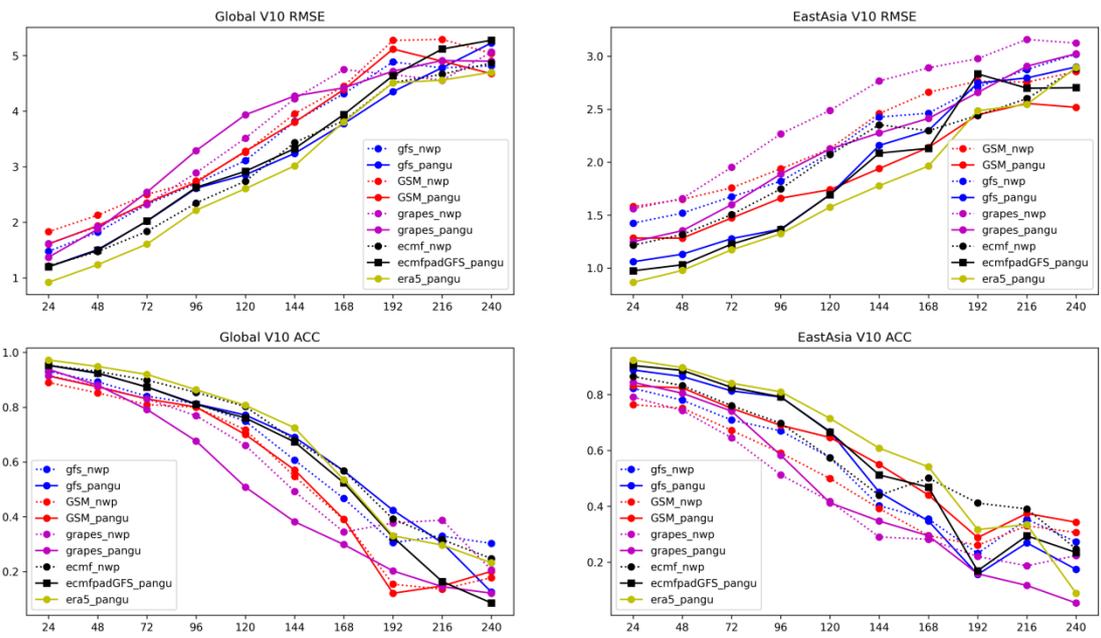

Figure A3.4 Same as Figure A3.1, but for V10.

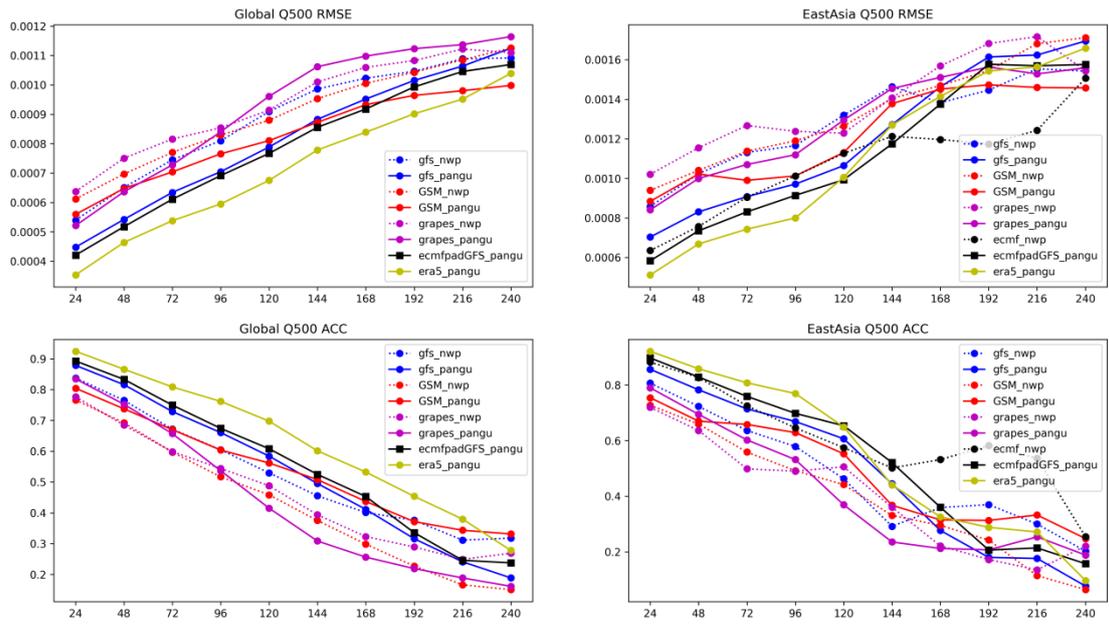

Figure A3.5 Same as Figure A3.1, but for Q500.

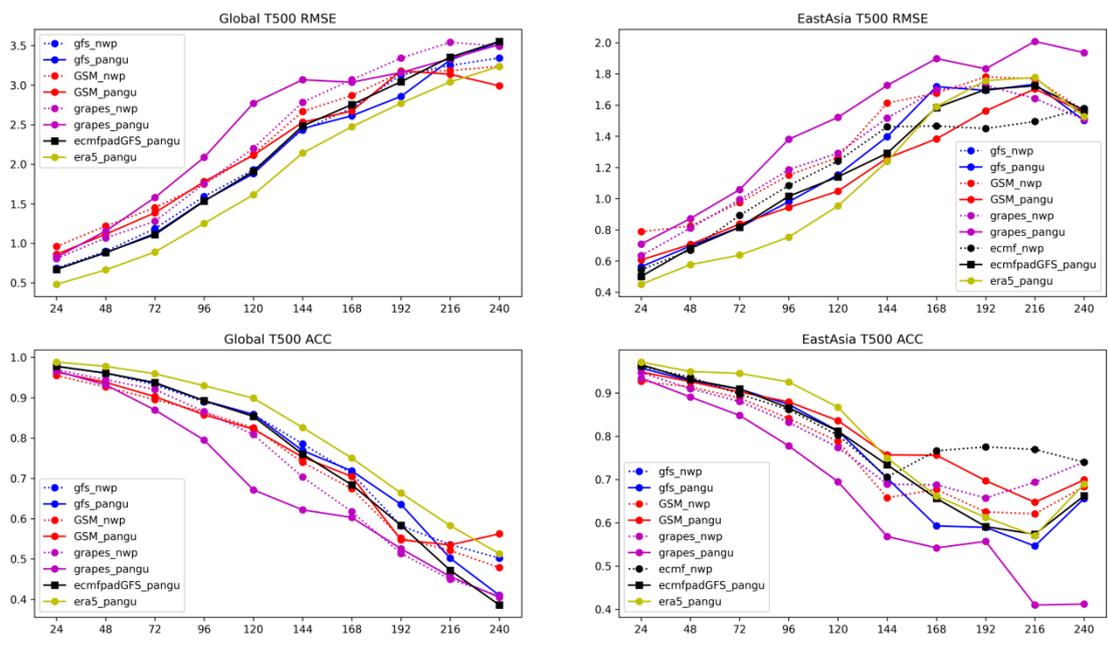

Figure A3.6 Same as Figure A3.1, but for T500.

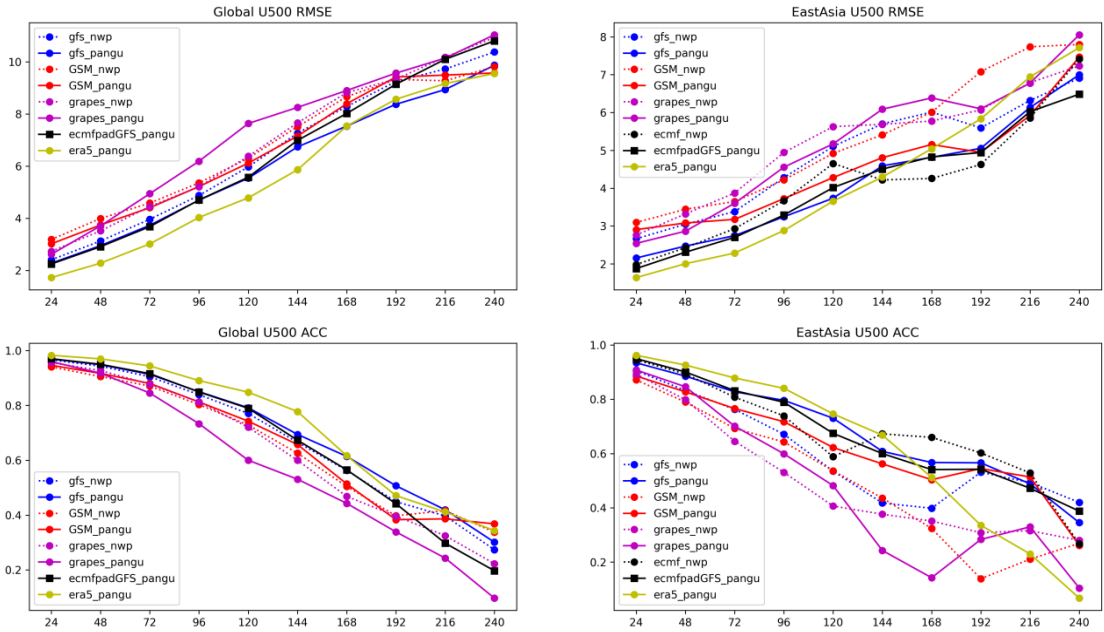

Figure A3.7 Same as Figure A3.1, but for U500.

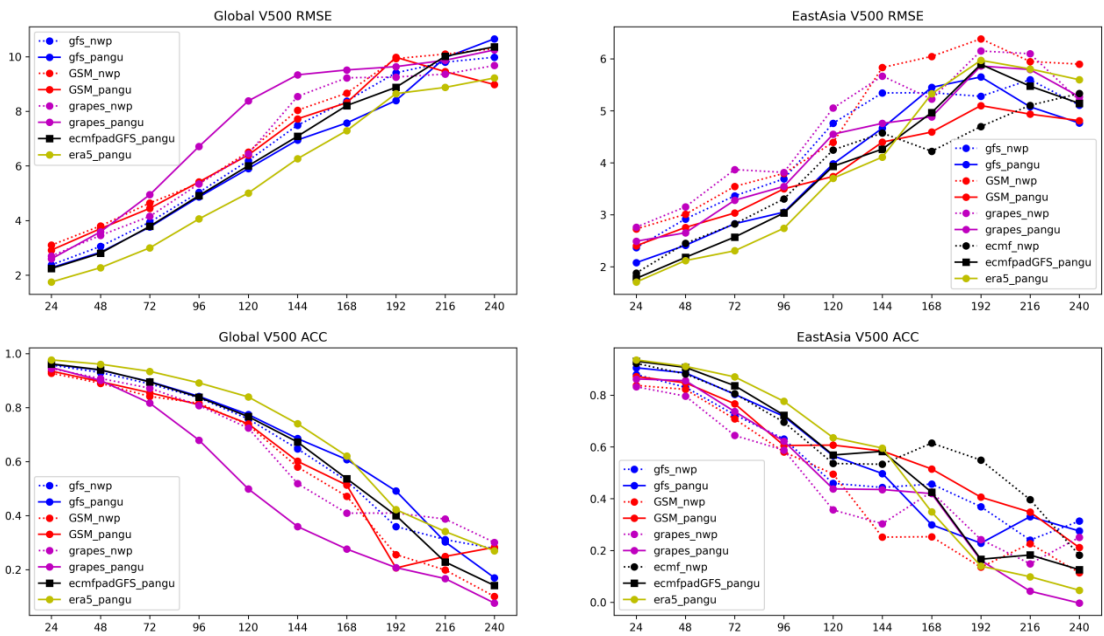

Figure A3.8 Same as Figure A3.1, but for V500.

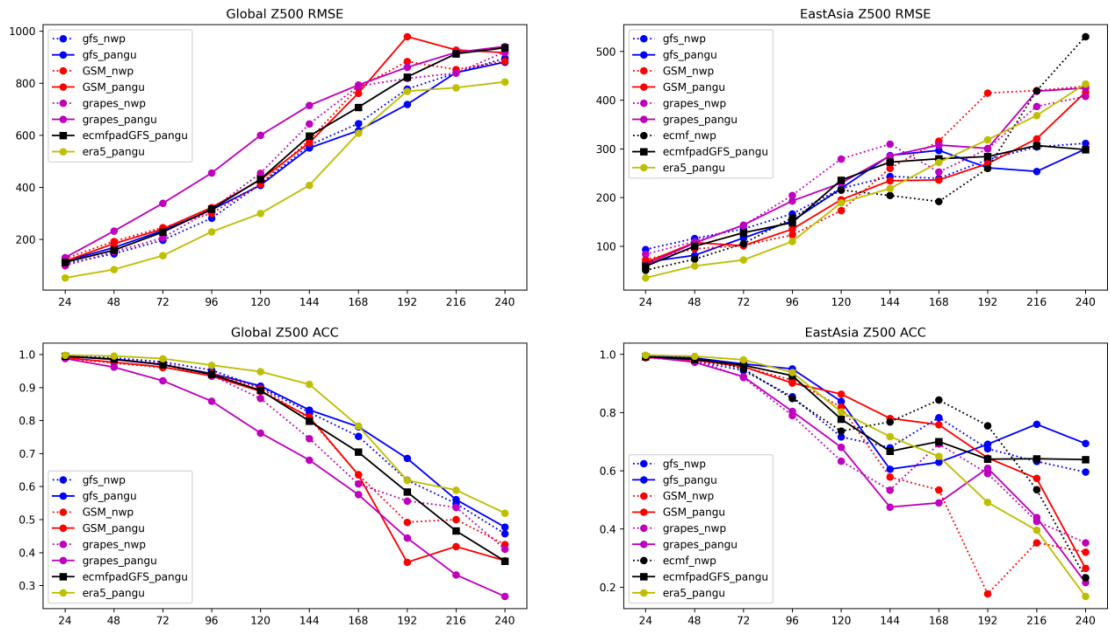

Figure A3.9 Same as Figure A3.1, but for Z500.